%% file: neurips_2021.tex
\documentclass{article}

\usepackage[nonatbib,final]{neurips_2021}

\usepackage[utf8]{inputenc} 
\usepackage[T1]{fontenc}    
\usepackage{hyperref}       
\usepackage{url}            
\usepackage{booktabs}       
\usepackage{amsfonts}       
\usepackage{nicefrac}       
\usepackage{microtype}      
\usepackage{xcolor}         
\usepackage{graphicx}
\usepackage{subcaption}
\usepackage{paralist}
\usepackage{wrapfig}

\usepackage{algorithm}
\usepackage{algorithmic}
\usepackage[algo2e,vlined,linesnumbered,ruled,resetcount]{algorithm2e}

\usepackage{amssymb}
\usepackage{amsthm}
\usepackage{amsmath}
\usepackage{multirow}
\usepackage{bbm}

\def\R{\mathbb{R}}
\def\IC{\mathrm{IC}}

\def \our{Zero Time Waste}
\def \ourabb{ZTW}

\usepackage{todonotes}

\newcommand{\stck}[1]{\textcolor{black}{#1}}

\title{Zero Time Waste: Recycling Predictions\\in Early Exit Neural Networks}

\author{
    Maciej Wo\l{}czyk\thanks{equal contribution}\;\;\thanks{Corresponding author: \texttt{maciej.wolczyk@doctoral.uj.edu.pl}} \\
    Jagiellonian University\\
    
    \And
    Bartosz Wójcik$^*$ \\
    Jagiellonian University\\
    
    \AND
    
    Klaudia Bałazy \\
    Jagiellonian University\\
    
    \And
    
    Igor Podolak \\
    Jagiellonian University\\
    
    \And 
    Jacek Tabor \\
    Jagiellonian University\\
    
    \And 
    Marek Śmieja \\
    Jagiellonian University\\
    
    \And 
    Tomasz Trzciński \\
    Jagiellonian University, \\
    Warsaw University of Technology, \\
    Tooploox
}

\usepackage[compact]{titlesec}
\usepackage{paralist}

\begin{document}


\maketitle

\begin{abstract}
The problem of reducing processing time of large deep learning models is a~fundamental challenge in many real-world applications. 
Early exit methods strive towards this goal by attaching additional Internal Classifiers ($\IC$s) to intermediate layers of a~neural network. $\IC$s can quickly return predictions for easy examples and, as a~result, reduce the average inference time of the whole model.
However, if a~particular $\IC$ does not decide to return an answer early, its predictions are discarded, with its computations effectively being wasted. 
To solve this issue, we introduce Zero Time Waste (ZTW), a~novel approach in which each $\IC$ reuses predictions returned by its predecessors by (1)~adding direct connections between $\IC$s and (2)~combining previous outputs in an ensemble-like manner.
We conduct extensive experiments across various datasets and architectures to demonstrate that ZTW achieves a~significantly better accuracy vs. inference~time trade-off than other recently proposed early exit methods.
\end{abstract}

\section{Introduction}

\begin{figure}[ht]
    \centering
    \begin{subfigure}[b]{0.57\textwidth}
        \includegraphics[width=\textwidth]{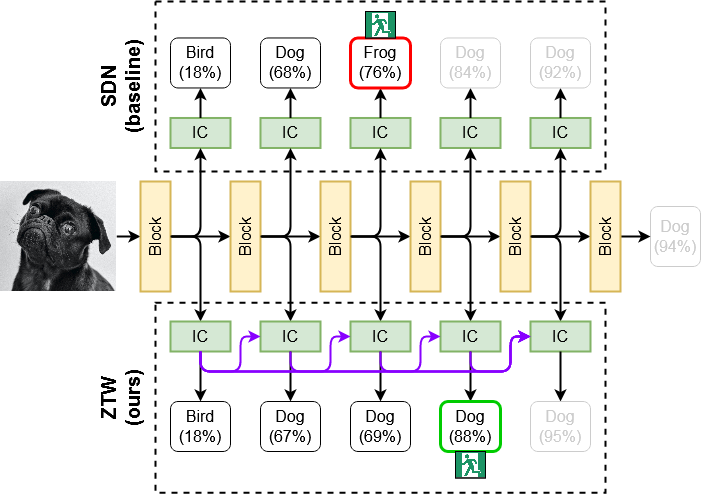}
        \caption{Comparison of the proposed ZTW (bottom) with a~conventional early-exit model, SDN (top).}
    \end{subfigure}
    \hfill
    \begin{subfigure}[b]{0.42\textwidth}
        \includegraphics[width=\textwidth]{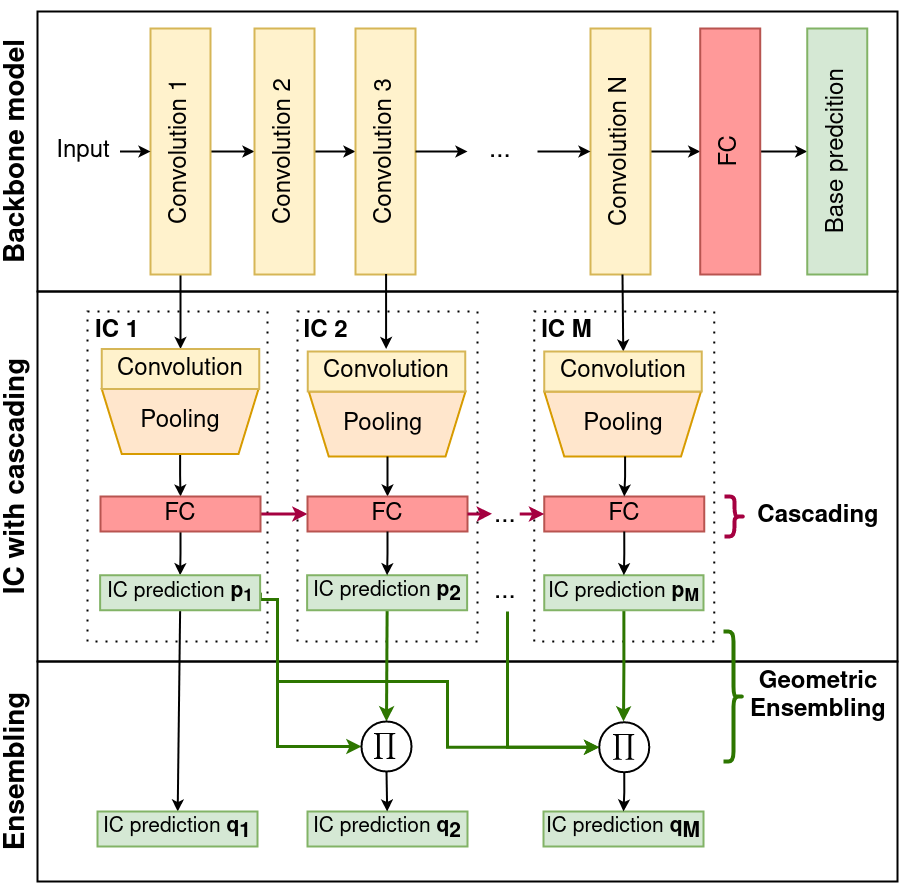}
        \caption{Detailed scheme of the proposed ZTW model architecture.}
    \end{subfigure}
    
    \caption{(a) In both approaches, internal classifiers ($\IC$s) attached to the intermediate hidden layers of the base network allow us to return predictions quickly for examples that are easy to process. While SDN discards predictions of uncertain $\IC$s ({\it e.g.} below a threshold of 75\%), ZTW reuses computations from all previous $\IC$s, which prevents information loss and waste of computational resources. (b)  Backbone network $f_\theta$ lends its hidden layer activations to $\IC$s, which share inferred information using \stck{cascade connections} (red horizontal arrows in the middle row) and give predictions $p_m$. The inferred predictions are combined using ensembling (bottom row) giving $q_m$.}
    
    \label{fig:re_intro}
\end{figure}

Deep learning models achieve tremendous successes across a~multitude of tasks, yet their training and inference often yield high computational costs and long processing times~\cite{he2016deep,krizhevsky2017imagenet}.
For some applications, however, efficiency remains a~critical challenge, {\it e.g.} to deploy a~reinforcement learning (RL) system in production the policy inference must be done in real-time~\cite{DBLP:journals/corr/abs-1904-12901}, while the robot performances suffer from the delay between measuring a~system state and acting upon it~\cite{Schuitema2010ControlDI}. Similarly, long inference latency in autonomous cars could impact its ability to control the speed~\cite{MLJ12-hester} and lead to accidents~\cite{grigorescu2020survey, jung2018perception}.

Typical approaches to reducing the processing complexity of neural networks in latency-critical applications include compressing the model~\cite{lee2021gst,livne2020pops,8852451} or approximating its responses~\cite{kouris2019approximate}. For instance, Livne \& Cohen~\cite{livne2020pops} propose to compress a~RL model by policy pruning, while Kouris et~al.~\cite{kouris2019approximate} approximate the responses of LSTM-based modules in self-driving cars to accelerate their inference time. While those methods improve processing efficiency, they still require samples to pass through the entire model. In contrast, biological neural networks leverage simple heuristics to speed up decision making, {\it e.g.} by shortening the processing path even in case of complex tasks~\cite{arielynorton2011fromthinkingtoolittle,gigerenzer2011heuristicdecisionmaking,kahneman2017thinking}.

This observation led a~way to the inception of the so-called {\it early exit} methods, such as Shallow-Deep Networks (SDN)~\cite{kaya2019shallow} and Patience-based Early Exit (PBEE)~\cite{zhou2020bert}, that attach simple classification heads, called internal classifiers~($\IC$s), to selected hidden layers of neural models to shorten the processing time. If the prediction confidence of a~given $\IC$ is sufficiently high, the response is returned, otherwise, the example is passed to the subsequent classifier. 
Although these models achieve promising results, they discard the response returned by early $\IC$s in the evaluation of the next $\IC$, disregarding potentially valuable information, e.g. decision confidence, and wasting computational effort already incurred.

Motivated by the above observation, we postulate to look at the problem of neural model processing efficiency from the {\it information recycling} perspective and introduce a~new family of {\it zero waste models}. More specifically, we investigate how information available at different layers of neural models can contribute to the decision process of the entire model. To that end, we propose \our{} (\ourabb{}), a~method for an intelligent aggregation of the information from previous $\IC$s. A~high-level view of our model is given in Figure~\ref{fig:re_intro}. Our approach relies on combining ideas from networks with skip connections~\cite{wang2018skipnet}, gradient boosting~\cite{bentejac2020comparative}, and ensemble learning~\cite{fort2019deep,lakshminarayanan2017simple}. Skip connections between subsequent $\IC$s (which we call \emph{cascade connections}) allow us to explicitly pass the information contained within low-level features to the deeper classifier, which forms a~\stck{cascading} structure of $\IC$s. In consequence, each $\IC$ improves on the prediction of previous $\IC$s, as in gradient boosting, instead of generating them from scratch. To give the opportunity for every $\IC$ to explicitly reuse predictions of all previous $\IC$s, we additionally build an ensemble of shallow $\IC$s.

We evaluate our approach on standard classification benchmarks, such as CIFAR-100 and ImageNet, as well as on the more latency-critical applications, such as reinforcement-learned models for interacting with sequential environments. To the best of our knowledge, we are the first to show that early exit methods can be used for cutting computational waste in a~reinforcement learning setting. 

Results show that \ourabb{} is able to save much more computation while preserving accuracy than current state-of-the-art early exit methods. In order to better understand where the improvements come from, we introduce Hindsight Improvability, a metric for measuring how efficiently the model reuses information from the past. We provide ablation studies and additional analysis of the proposed method in the Appendix. 

To summarize, the contributions of our work are the following:
\begin{itemize}
\item We introduce a family of zero waste models that quantify neural network efficiency with the Hindsight Improvability metrics.
\item We propose an instance of zero waste models dubbed Zero Time Waste (ZTW) method which uses cascade connections and ensembling to reuse the responses of previous ICs for the final decision.
\item We show how the state-of-the-art performance of ZTW in the supervised learning scenario generalizes to reinforcement learning.
\end{itemize}

\section{Related Work}

The drive towards reducing computational waste in deep learning literature has so far focused on reducing the inference time. Numerous approaches for accelerating deep learning models focus on building more efficient architectures~\cite{howard2017mobilenets}, reducing the number of parameters~\cite{he2017channel} or distilling knowledge to smaller networks~\cite{44873}. Thus, they decrease inference time by reducing the overall complexity of the model instead of using the conditional computation framework of adapting computational effort to each example. As such we find them orthogonal to the main ideas of our work, e.g. we show that applying our method to architectures designed for efficiency, such as MobileNet \cite{howard2017mobilenets}, leads to even further acceleration. Hence, we focus here on methods that adaptively set the inference time for each example.

\paragraph{Conditional Computation}

Conditional computation was first proposed for deep neural networks in Bengio et~al.~\cite{bengio2013estimating} and Davis~\&~Arel~\cite{davis2013low}, and since then many sophisticated methods have been proposed in this field, including dynamic routing \cite{mcgill2017deciding}, cascading with multiple networks \cite{wang2017idk} and skipping intermediate layers~\cite{wang2018skipnet} or channels~\cite{wang2020deep}. In this work, we focus on the family of early exit approaches, as they usually do not require special assumptions about the underlying architecture of the network and the training paradigm, and because of that can be easily applied to many commonly used architectures. In BranchyNet~\cite{teerapittayanon2016branchynet} a~loss function consisting of a~weighted sum of individual head losses is utilized in training, and entropy of the head prediction is used for the early exit criterion. Berestizshevsky \& Guy \cite{berestizshevsky2019dynamically} propose to use confidence (maximum of the softmax output) instead. A broader overview of early exit methods is available in Scardapane~et~al.~\cite{scardapane2020should}.

Several works proposed specialized architectures for conditional computation which allow for multi-scale feature processing~\cite{huang2018multi,yang2020resolution,wang2020glance}, and developed techniques to train them more efficiently by passing information through the network~\cite{phuong2019distillation,li2019improved}. However, in this paper, we consider the case of increasing inference speed of a pre-trained network based on an architecture which was not built with conditional computation or even efficiency in mind. We argue that this is a practical use case, as this approach can be used to a wider array of models. As such, we do not compare with these methods directly.

Shallow-Deep Networks (SDN) \cite{kaya2019shallow} is a~conceptually simple yet effective method, where the comparison of confidence with a~fixed threshold is used as the exit criterion. The authors attach internal classifiers to layers selected based on the number of compute operations needed to reach them. 
The answer of each head is independent of the answers of the previous heads, although in a separate experiment the authors analyze the measure of disagreement between the predictions of final and intermediate heads. 

Zhou et al. \cite{zhou2020bert} propose Patience-based Early Exit (PBEE) method, which terminates inference after $t$ consecutive unchanged answers, and show that it outperforms SDN on a range of NLP tasks.
The idea of checking for agreement in preceding $\IC$s is connected to our approach of reusing information from the past. However, we find that applying PBEE in our setting does not always work better than SDN. Additionally, in the experiments from the original work, PBEE was trained simultaneously along with the base network, thus making it impossible to preserve the original pre-trained model.

\paragraph{Ensembles}
Ensembling is typically used to improve the accuracy of machine learning models~\cite{dietterich2000ensemble}. Lakshminarayanan et al.~\cite{lakshminarayanan2017simple} showed that it also greatly improves calibration of deep neural networks. There were several attempts to create an ensemble from different layers of a~network. Scardapane et al.~\cite{scardapane2020differentiable} adaptively exploit outputs of all internal classifiers, albeit not in a~conditional computation context. Phuong \& Lampert~\cite{phuong2019distillation} used averaged answers of heads up to the current head for anytime-prediction, where the computational budget is unknown. Besides the method being much more basic, their setup is notably different from ours, as it assumes the same computational budget for all samples no matter how difficult the example is. Finally, none of the ensemble methods mentioned above were designed to work with pre-trained models.

\section{Zero Time Waste}

Our goal is to reduce computational costs of neural networks by minimizing redundant operations and information loss. To achieve it, we use the conditional computation setting, in which we dynamically select the route of an input example in a~neural network. By controlling the computational route, we can decide how the information is stored and utilized within the model for each particular example. Intuitively, difficult examples require more resources to process, but using the same amount of compute for easy examples is wasteful. Below we describe our Zero Time Waste method in detail.

In order to adapt already trained models to conditional computation setting, we attach and train early exit classifier heads on top of several selected layers, without changing the parameters of the base network.
During inference, the whole model exits through one of them when the response is likely enough, thus saving computational resources.

Formally, we consider a~multi-class classification problem, where $x \in \R^D$ denotes an input example and $y \in \{1,\ldots,K\}$ is its target class. Let $f_\theta: \R^D \to \R^K$ be a pre-trained neural network with logit output designed for solving the above classification task.  The weights $\theta$ will not be modified.

\begin{figure}
    \centering
    \label{fig:model_graph}
\end{figure}

\paragraph{Model overview} 

Following typical early exit frameworks, we add $M$ shallow Internal Classifiers, $\IC_1,\ldots,\IC_M$, on intermediate layers of $f_\theta$. Namely, let $g_{\phi_m}$, for $m\in\{1,\ldots,M\}$, be the $m$-th $\IC$ network returning $K$ logits, which is attached to hidden layer $f_{\theta_{m}}$ of the base network $f_\theta$. The index $m$ is independent of $f_\theta$ layer numbering. In general, $M$ is lower than the overall number of $f_\theta$ hidden layers since we do not add $\IC$s after every layer (see more details in Appendix \ref{sec:appendix_placement_ic}). 

Although using $\IC$s to return an answer early can reduce overall computation time~\cite{kaya2019shallow}, in a~standard setting each $\IC$ makes its decision independently, ignoring the responses returned by previous $\IC$s. As we show in Section~\ref{sec:information_loss}, early layers often give correct answers for examples that are misclassified by later classifiers, and hence discarding their information leads to waste and performance drops. 
To address this issue, we need mechanisms that collect the information from the first $(m-1)$ $\IC$s to inform the decision of $\IC_m$. 
For this purpose, we introduce two complementary techniques: \emph{cascade connections} and \emph{ensembling}, and show how they help reduce information waste and, in turn, accelerate the model.

\stck{Cascade connections} directly transfer the already inferred information between consecutive $\IC$s instead of re-computing it again. 
Thus, they improve the performance of initial $\IC$s that lack enough predictive power to classify correctly based on low-level features.
Ensembling of individual $\IC$s improves performance as the number of members increases, thus showing greatest improvements in the deeper part of the network. This is visualized in Figure~\ref{fig:re_intro} where \stck{cascade connections} are used first to pass already inferred information to later $\IC$s, while ensembling is utilized to conclude the $\IC$ prediction. The details on these two techniques are presented in the following paragraphs.

\paragraph{\stck{Cascade connections}}  
Inspired by the gradient boosting algorithm and literature on cascading classifiers \cite{viola2004robust}, we allow each $\IC$ to improve on the predictions of previous $\IC$s instead of inferring them from scratch. The idea of \stck{cascade connections} is implemented by adding skip connections that combine the output of the base model hidden layer $f_{\theta_m}$ with the logits of $\IC_{m-1}$ and pass it to $\IC_m$. The prediction is realized by the softmax function applied to $g_{\phi_m}$ (the $m$-th $\IC$ network):
\begin{equation}
\label{eq:stacking}
p_m = 
\mathrm{softmax}(g_{\phi_m} (f_{\theta_m}(x), g_{\phi_{m-1}} \circ f_{\theta_{m-1}}(x))) \text{, for } m > 1,
\end{equation}
where $g \circ f (x) = g(f(x))$ denotes the composition of functions. Formally, $p_m = p_m(x;\phi_m)$, where $\phi_m$ are trainable parameters of $\IC_m$, but we drop these parameters in notation for brevity. $\IC_1$ uses only the information coming from the layer $f_{\theta_1}$ which does not need to be the first hidden layer of $f_\theta$. Figure~\ref{fig:re_intro} shows the skip connections as red horizontal arrows. 

Each $\IC_m$ is trained in parallel (with respect to $\phi_m$) to optimize the prediction of all output classes using an appropriate loss function $\mathcal{L}(p_m)$, e.g. cross-entropy for classification. However, during the backward step it is crucial to stop the gradient of a~loss function from passing to the previous classifier. Allowing the gradients of loss $\mathcal{L}(p_{m})$ to affect $\phi_{j}$ for $j \in {1, .., m-1}$  leads to a~significant performance degradation of earlier layers due to increased focus on the features important for $\IC_m$, as we show in Appendix~\ref{sec:stop_gradient}.

\paragraph{Ensembling} 
Ensembling in machine learning models reliably increases the performance of a model while improving robustness and uncertainty estimation~\cite{fort2019deep,lakshminarayanan2017simple}. The main drawback of this approach is its wastefulness, as it requires to train multiple models and use them to process the same examples. However, in our setup we can adopt this idea to combine predictions which were already pre-computed in previous $\IC$s, with near-zero additional computational cost. 

To obtain a~reliable zero-waste system, we build ensembles that combine outputs from groups of $\IC$s to provide the final answer of the $m$-th classifier. Since the classifiers we are using vary significantly in predictive strength (later $\IC$s achieve better performance than early $\IC$s) and their predictions are correlated, the standard approach to deep model ensembling does not work in our case.
Thus, we introduce weighted geometric mean with class balancing, which allows us to reliably find a~combination of pre-computed responses that maximizes the expected result.

Let $p_1,p_2,\dots,p_m$ be the outputs of $m$ consecutive $\IC$ predictions (after \stck{cascade connections stage}) for a~given $x$ (Figure~\ref{fig:re_intro}). We define the probability of the $i$-th class in the $m$-th ensemble to be:
\begin{equation}
\label{eq:ensemble_eq}
q_m^i(x){}=\frac{1}{Z_m}\,{}b_m^i\prod_{j\leq{}m}\big(p_j^i(x)\big)^{w_m^j},
\end{equation}
where $b_m^i > 0$ and $w_m^j>0$, for $j=1,\dots,m$, are trainable parameters, and $Z_m$ is a~normalization factor, such that $\sum_i q_m^i(x) = 1$. Observe that $w_m^j$ can be interpreted as our prior belief in predictions of $\IC_j$, i.e. large weight $w_m^j$ indicates less confidence in the predictions of $\IC_j$. On the other hand, $b_m^i$ represents the prior of $i$-th class for $\IC_m$. The $m$ indices in $w_m$ and $b_m$ are needed as the weights are trained independently for each subset $\{\IC_j: j\leq m\}$. Although there are viable potential approaches to setting these parameters by hand, we verified that optimizing them directly by minimizing the cross-entropy loss on the training dataset works best.

Out of additive and geometric ensemble settings we found the latter to be preferable. In this formulation, a~low class confidence of a~single $\IC$ would significantly reduce the probability of that class in the whole ensemble. In consequence, in order for the confidence of the given class to be high, we require all $\IC$s to be confident in that class. Thus, in geometric ensembling, an incorrect although confident $\IC$ answer has less chance of ending calculations prematurely. In the additive setting,
the negative impact of a~single confident but incorrect $\IC$ is much higher, as we show in Appendix~\ref{sec:additive_vs_geometric}. Hence our choice of geometric ensembling.

Direct calculation of the product in~\eqref{eq:ensemble_eq} might lead to numerical instabilities whenever the probabilities are close to zero. To avoid this problem we note that
\begin{equation*}
b_m^i\prod_{j\leq{}m}\big(p_j^{i}(x)\big)^{w_m^j}=b_m^i\exp\bigg(\sum_{j\leq{}m} w_m^j \ln{}p_j^{i}(x)\bigg)
\label{eq:expln},
\end{equation*}
and that log-probabilities $\ln p_j^{i}$ can be obtained by running the numerically stable log~softmax function on the logits $g_{\phi_m}$ of the classifier.

\begin{algorithm}[tb]
   \caption{Zero Time Waste}
  \SetKwFor{StackFor}{For}{do in parallel {\rm$\enskip \quad \quad \quad \triangleright$ \stck{Cascade connection} $\IC$s}}{}
  \SetKwFor{EnsembleFor}{For}{do {\rm$\quad \quad \quad \quad \triangleright$ Geometric Ensembling}}{}

   {\bfseries Input:} pre-trained model $f_\theta$, cross-entropy loss function $\mathcal{L}$, training set $\mathcal{T}$.\;
   
   {\bfseries Initialize} $M$ shallow models $g_{\phi_m}$ at selected layers $f_{\theta_m}$.\;
   
   \StackFor{$m=1,\dots,M$}{ 
        Set $p_m$ according to~\eqref{eq:stacking}.\;
   
        minimize $\mathbb{E}_{(x, y) \in \mathcal{T}} \left [ \mathcal{L}(p_{m}(x), y) \right ]$ wrt. $\phi_m$ by gradient descent\;
    }
    
   \EnsembleFor{$m=1,\dots,M$}{
       Initialize $w_m, b_m$ and define $q_m(x){}$ according to~\eqref{eq:ensemble_eq}.\;
       
       minimize $\mathbb{E}_{(x, y) \in \mathcal{T}} \left [ \mathcal{L}(q_m(x), y) \right ] $ wrt. $w_m, b_m$ by gradient descent\;
    }
\label{alg:ztw}
\end{algorithm}

Both cascade connections and ensembling have different impact on the model. Cascade connections primarily boost the accuracy of early $\IC$s.
Ensembling, on the other hand, improves primarily the performance of later $\IC$s, which combine the information from many previous classifiers.

This is not surprising, given that the power of the ensemble increases with the number of members, provided they are at least weak in the sense of boosting theory~\cite{schapire_strength_1990}. 
As such, the two techniques introduced above are complementary, which we also show empirically via ablation studies in Appendix \ref{sec:ablation}. The whole training procedure is presented in Algorithm~\ref{alg:ztw}.

\paragraph{Conditional inference} Once a~\ourabb{} model is trained, the~following question appears: how to use the constructed system at test time? More precisely, we need to dynamically find the shortest processing path for a~given input example. For this purpose, we use one of the standard confidence scores given by the probability of the most confident class. If the $m$-th classifier is confident enough about its prediction, i.e. if
\begin{equation}
\underset{i}{\max}\, q_m^i > \tau \text{, for a~fixed } \tau>0,
\label{eq:thres}
\end{equation}
where $i$ is the~class index, then we terminate the computation and return the response given by this $\IC$. If this condition is not satisfied, we continue processing $x$ and go to the next $\IC$.

Threshold $\tau$ in~\eqref{eq:thres} is a~manually selected value, which controls the acceleration-performance trade-off of the model. A~lower threshold leads to a~significant speed-up at the cost of a~possible drop in accuracy. Observe that for $\tau>1$, we recover the original model $f_\theta$, since none of the $\IC$s is confident enough to answer earlier. In practice, to select its appropriate value, we advise using a~held-out set to evaluate a~range of possible values of $\tau$.

\section{Experiments}
In this section we examine the performance of \our{} and analyze its impact on waste reduction in comparison to two recently proposed early-exit methods: (1)~Shallow-Deep Networks (SDN)~\cite{kaya2019shallow} and (2)~Patience-Based Early Exit (PBEE)~\cite{zhou2020bert}. In contrast to SDN and PBEE, which train $\IC$s independently, \ourabb{} reuses information from past classifiers to improve the performance. SDN and \ourabb{} use maximum class probability as the confidence estimator, while PBEE checks the number of classifiers in sequence that gave the same prediction. For example, for PBEE $\tau = 2$ means that if the answer of the current $\IC$ is the same as the answers of the $2$ preceding $\IC$s, we can return that answer, otherwise we continue the computation.

In our experiments, we measure how much computation we can save by re-using responses of $\IC$s while keeping good performance, hence obeying the zero waste paradigm. To evaluate the efficiency of the model, we compute the average number of floating-point operations required to perform the forward pass for a~single sample. We use it as a hardware-agnostic measure of inference cost and refer to it simply as the "inference time" in all subsequent references.
For the evaluation in supervised learning, we use three datasets: CIFAR-10, CIFAR-100, and Tiny~ImageNet, and four commonly used architectures: ResNet-56~\cite{he2016deep}, MobileNet~\cite{howard2017mobilenets}, WideResNet~\cite{zagoruyko2016wide}, and VGG-16BN~\cite{DBLP:journals/corr/SimonyanZ14a} as base networks. We check all combinations of methods, datasets, and architectures, giving $3 \cdot 3 \cdot 4 = 36$ models in total, and we additionally evaluate a~single architecture on the ImageNet dataset to show that the approach is scalable. Additionally, we examine how \our{} performs at reducing waste in a~reinforcement learning setting of Atari 2600 environments. To the best of our knowledge, we are the first to apply early exit methods to reinforcement learning.

Appendix~\ref{sec:training_details} describes the details about the network architecture, hyperparameters, and training process. Appendix~\ref{sec:all_plots} contains extended plots and tables, and results of an additional transfer learning experiment. In Appendix~\ref{sec:ablation} we provide ablation studies, focusing in particular on analyzing how each of the proposed improvements affects the performance, and empirically justifying some of the design choices (e.g. geometric ensembles vs. additive ensembles). We provide the source code for our experiments at \url{https://github.com/gmum/Zero-Time-Waste}.

\input{results_table_single}

\subsection{Time Savings in Supervised Learning}

We check what percentage of computation of the base network can be saved by reusing the information from previous layers in a~supervised learning setting. To do this, we evaluate how each method behaves at a~particular fraction of the computational power (measured in floating point operations) of the base network. We select the highest threshold $\tau$ such that the average inference time is smaller than, for example, $25\%$ of the original time. Then we calculate accuracy for that threshold. Table~\ref{tab:results_table} contains summary of this analysis, averaged over three seeds, with further details (plots for all thresholds, standard deviations) shown in Appendix~\ref{sec:additional_results_supervised}. 

Looking at the results, we highlight the fact that methods which do not reuse information between $\IC$s do not always achieve the goal of reducing computational waste. For example, SDN and PBEE cannot maintain the accuracy of the base network for MobileNet on Tiny ImageNet when using the same computational power, scoring respectively $0.4$ and $3.7$ percentage points lower than the baseline. Adding $\IC$s to the network and then discarding their predictions when they are not confident enough to return the final answer introduces computational overhead without any gains.  By reusing the information from previous $\IC$s \ourabb{} overcomes this issue and maintains the accuracy of the base network for all considered settings. In particular cases, such as ResNet-56 on Tiny ImageNet or MobileNet on Cifar-100, \our{} even significantly outperforms the core network.

Similar observation can be made for other inference time limits as well. \ourabb{} consistently maintains high accuracy using less computational resources than the other approaches, for all combinations of datasets and architectures. Although PBEE reuses information from previous layers to decide whether to stop computation or not, this is not sufficient to reduce the waste in the network. While PBEE outperforms SDN when given higher inference time limits, it often fails for smaller limits ($25\%, 50\%$). We hypothesize that this is result of the fact that PBEE has smaller flexibility with respect to $\tau$. While for SDN and \ourabb{} values of $\tau$ are continuous, for PBEE they represent a~discrete number of $\IC$s that must sequentially agree before returning an answer.

Finally, we check whether our observations scale up to larger datasets by running experiments on ImageNet using a pre-trained ResNet-50 from the torchvision package\footnote{\url{https://pytorch.org/vision/stable/index.html}}. The results presented in Table \ref{tab:imagenet_results} show that \our{} is able to gain significant improvements over the two tested baselines even in this more challenging setting. Additional details of this experiments are presented in Appendix \ref{sec:imagenet_results}.

\begin{wraptable}{r}{0.5\textwidth}
    \caption{ImageNet results (test accuracy in percentage points) show that zero-waste approach scales up to larger datasets.}
    \label{tab:imagenet_results}
    \small
    \centering
    \begin{tabular}{crrrr}
        \toprule
        Algo & 25\% & 50\% & 75\% & 100\% \\
        \midrule
        SDN & 33.8 & 53.8 & 69.7 & 75.8 \\
        PBEE & 28.3 & 28.3 & 62.9 & 73.3 \\
        ZTW & \textbf{34.9} &\textbf{54.9} & \textbf{70.6} & \textbf{76.3} \\
        \bottomrule
    \end{tabular}
\end{wraptable}

Given the performance of \ourabb{}, the results show that paying attention to the minimization of computational waste leads to tangible, practical improvements of the inference time of the network. Therefore, we devote next section to explaining where the empirical gains come from and how to measure information loss in the models.

\subsection{Information Loss in Early Exit Models}
\label{sec:information_loss}

\begin{figure}
    \centering
    \includegraphics[width=0.495\textwidth]{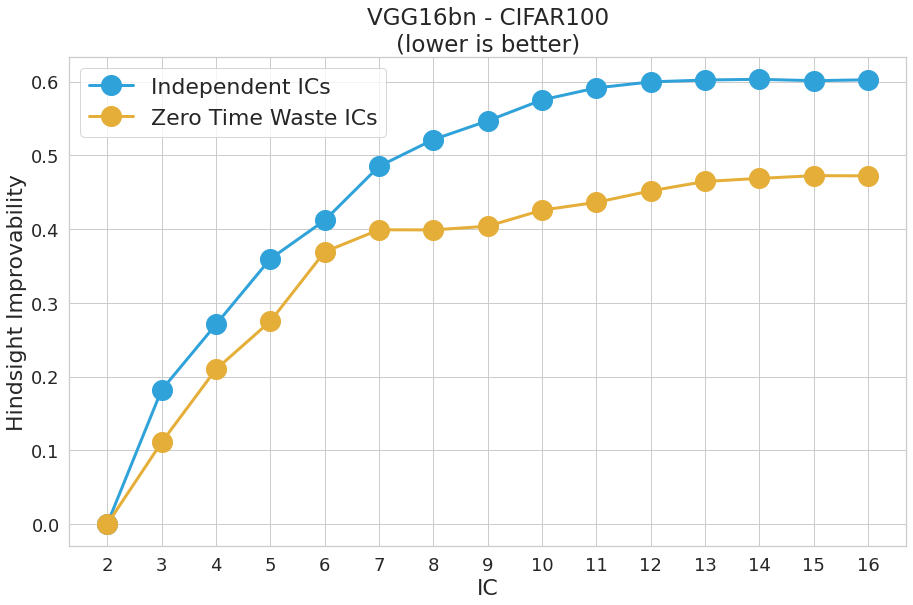}
    \includegraphics[width=0.495\textwidth]{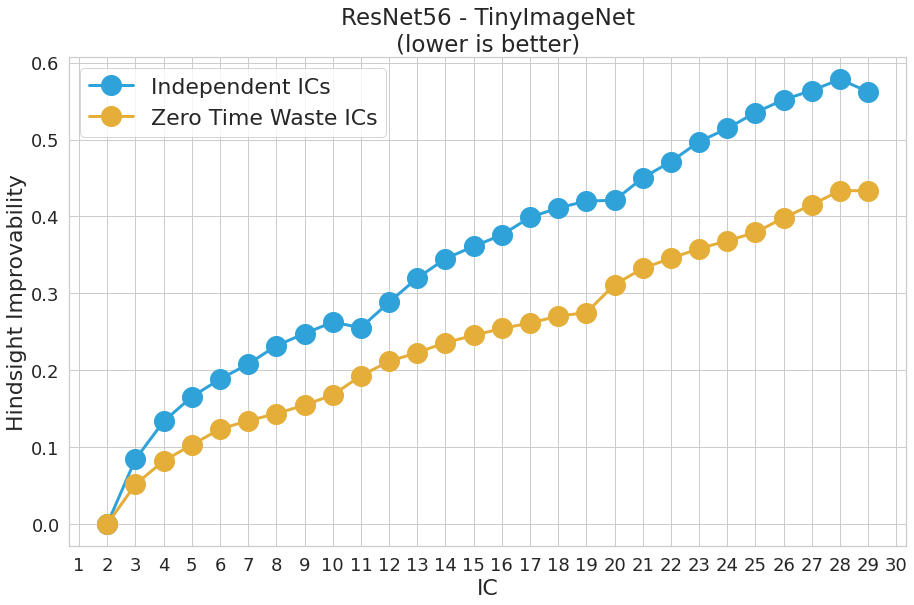}
    \caption{Hindsight Improvability. For each $\IC$ (horizontal axis) we look at examples it misclassified and we check how many of them were classified correctly by any of the previous $\IC$s. The lower the number, the better the $\IC$ is at reusing previous information.}
    \label{fig:hindsight_improv}
\end{figure}

Since $\IC$s in a~given model are heavily correlated, it is not immediately obvious why reusing past predictions should improve performance. Later $\IC$s operate on high-level features for which class separation is much easier than for early $\IC$s, and hence get better accuracy. Thus, we ask a question --- is there something that early $\IC$s know that the later $\IC$s do not?

For that purpose, we introduce a~metric to evaluate how much a~given $\IC$ could improve performance by reusing information from all previous $\IC$s. We measure it by checking how many examples incorrectly classified by $\IC_m$ were classified correctly by any of the previous $\IC$s.
An $\IC$ which reuses predictions from the past perfectly would achieve a~low score on this metric since it would remember all the correct answers of the previous $\IC$s. On the other hand, an $\IC$ in a~model which trains each classifier independently would have a~higher score on this metric, since it does not use past information at all.  We call this metric Hindsight Improvability (HI) since it measures how many mistakes we would be able to avoid if we used information from the past efficiently.

Let $\mathcal{C}_m$ denote the set of examples correctly classified by $\IC_m$, with its complement $\overline{\mathcal{C}}_m$ being the set of examples classified incorrectly. To measure the Hindsight Improvability of $\IC_m$ we calculate:

\begin{equation*}
    \text{HI}_m = \frac{\left |\overline{\mathcal{C}}_m \cap (\bigcup_{n < m} \mathcal{C}_n) \right |} {\left |\overline{\mathcal{C}}_m \right |}
\end{equation*}

Figure~\ref{fig:hindsight_improv} compares the values of HI for a~method with independent $\IC$s (SDN in this case) and \ourabb{} which explicitly recycles computations. In the case of VGG16 trained with independent $\IC$s, over $60\%$ of the mistakes could be avoided if we properly used information from the past, which would translate to improvement from $71.5\%$ to $82.9\%$ accuracy. Similarily, for ResNet-56 trained on TinyImageNet, the number of errors could be cut by around 57\%. 

\ourabb{} consistently outperforms the baseline, with the largest differences visible at the later $\IC$s, which can in principle gain the most from reusing previous predictions. Thus, \our{} is able to efficiently recycle information from the past. At the same time, there is still a~room for significant improvements, which shows that future zero waste approaches could offer additional enhancements. 

\subsection{Time Savings in Reinforcement Learning}

Although supervised learning is an important testbed for deep learning, it does not properly reflect the challenges encountered in the real world. In order to examine the impact of waste-minimization methods in a~setting that reflects the sequential nature of interacting with the world, we evaluate it in a~Reinforcement Learning (RL) setting. In particular, we use the environments from the suite of Atari 2600 games~\cite{mnih2015human}. 
\begin{figure}
    \centering
    \includegraphics[width=0.495\textwidth]{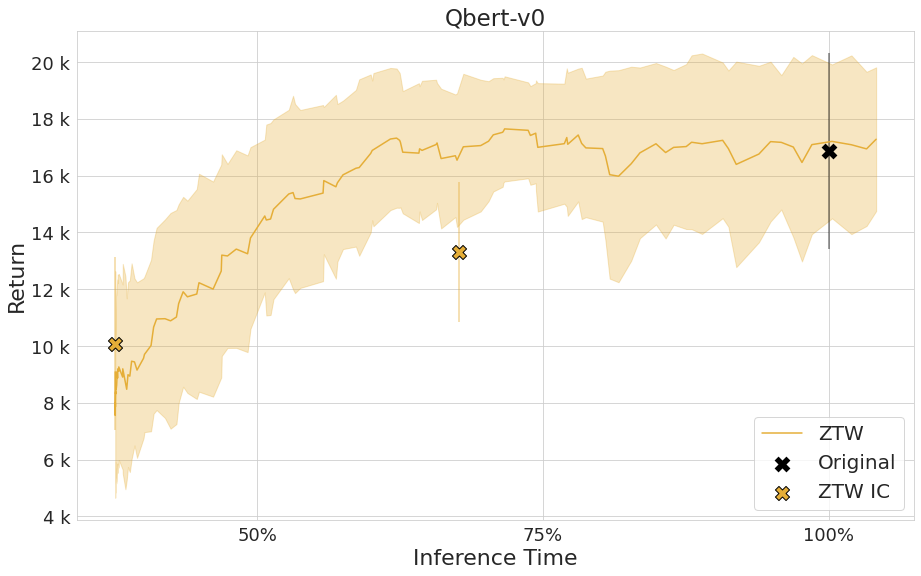}
    \includegraphics[width=0.495\textwidth]{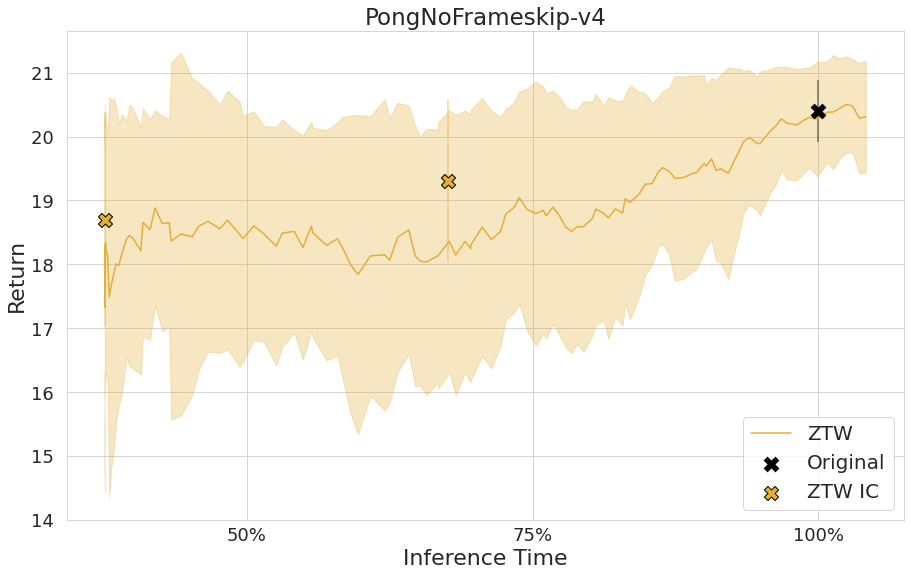}
    \caption{
    Inference time vs. average return of the ZTW policy in an~RL setting on Q*bert and Pong Atari 2600 environments. The plot was generated by using different values of the confidence threshold $\tau$ hyperparameter.  Since the RL environments are stochastic, we plot the return with a~standard deviation calculated on 10 runs. \ourabb{} saves a~significant amount of computation while preserving the original performance, showcasing that waste can be minimized also in the reinforcement learning domain.}
    \label{fig:rl_results}
\end{figure}

Similarly as in the supervised setting, we start with a~pre-trained network, which in this case represents a~policy trained with the Proximal Policy Optimization (PPO) algorithm~\cite{schulman2017proximal}. We attach the $\IC$s to the network and train it by distilling the knowledge from the core network to the $\IC$s. We use a~behavioral cloning approach, where the states are sampled from the policy defined by the $\IC$s and the labels are provided by the expert model. Since actions in Atari 2600 are discrete, we can then use the same confidence threshold-based approach to early exit inference as in the case of classification. 
More details about the training process are provided in the Appendix~\ref{sec:training_details_rl}.

In order to investigate the relationship between computation waste reduction and performance, we evaluate \our{} for different values of confidence threshold $\tau$. By setting a~very high $\tau$ value, we retrieve the performance of the original model (none of the $\IC$s respond) and by slowly decreasing its value we can reduce the computational cost ($\IC$s begin to return answers earlier).
In Figure~\ref{fig:rl_results} we check values of $\tau$ in the interval $[0.1, 1.0]$ to show how \ourabb{} is able to control the acceleration-performance balance for Q*Bert and Pong, two popular Atari 2600 environments.
By setting lower $\tau$ thresholds for Q*Bert we can save around $45\%$ of computations without score degradation. Similarly, for Pong we can get $60\%$ reduction with minor impact on performance (note that average human score is 9.3 points).  This shows that even the small four-layered convolutional architecture commonly used for Atari~\cite{mnih2015human} introduces a~noticeable waste of computation which can be mitigated within a~zero-waste paradigm. We highlight this fact as the field of reinforcement learning has largely focused on efficiency in terms of number of samples and training time, while paying less attention to the issue of efficient inference.

\subsection{Impact \& Limitations}
\label{sec:limitations}

Our framework is the cornerstone of an environmental-aware computation where information recycling within a~model is cautiously studied to avoid wasting resources. The focus on computational efficiency, however, introduces a~natural trade-off between model accuracy and its computational cost. Although in most cases we can carefully adjust the appropriate method hyperparameters to avoid a~significant accuracy drop, some testing samples remain surprisingly challenging for ZTW, which indicates a~need for further investigation of the accuracy vs. computation cost trade-off offered by our method.

\begin{figure}
    \centering
    \includegraphics[width=\textwidth]{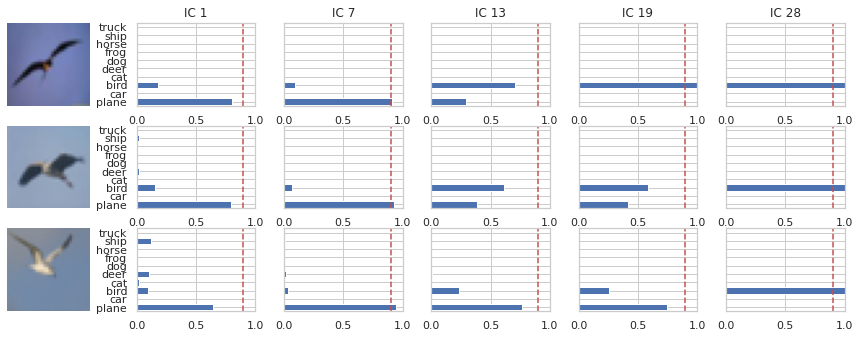}
    \caption{Examples of bird images which were incorrectly classified as airplanes by \ourabb{}. The early $\IC$s are misled by the low-level features (blue sky, sharp edges, grayscale silhouette) and return a~prediction before the later $\IC$s can detect more subtle high-level features.}
    \label{fig:failure_modes}
\end{figure}

Figure~\ref{fig:failure_modes} contains examples of images for which low-level features in a~given image consistently point at a~wrong class, while high-level features would allow us to deduce the correct class.  Images of birds which contain sharp lines and grayscale silhouettes are interpreted as airplanes by early $\IC$s which operate on low-level features. If the confidence of these classifiers gets high enough, the answer might be returned before later classifiers can correct this decision. We highlight the problem of dealing with examples which are seemingly easy but turn out difficult as an important future direction for conditional computation methods.

\section{Conclusion}
In this work, we show that discarding predictions of the previous $\IC$s in early exit models leads to waste of computation resources and a~significant loss of information. This result is supported by the introduced Hindsight Improvability metric, as well as empirical result for reducing computations in existing networks. The proposed \our{} method attempts to solve these issues by incorporating outputs from the past heads by using cascade connections and geometric ensembling.
We show that \ourabb{} outperforms other approaches on  multiple standard datasets and architectures for supervised learning, as well as in Atari 2600 reinforcement learning suite. At the same time we postulate that focusing on reducing the computational waste in a~safe and stable way is an important direction for future research in deep learning.

\section*{Acknowledgments \& Funding Disclosure}

This research was funded by Foundation for Polish Science (grant no POIR.04.04\allowbreak.00-00-14DE/18-00 carried out within the Team-Net program co-financed by the European Union under the European Regional Development Fund) and National Science Centre, Poland (grant no 2018/31/B/ST6/00993 and grant no 2020/39/B/ST6/01511). The authors have applied a CC BY license to any Author Accepted Manuscript (AAM) version arising from this submission, in accordance with the grants’ open access conditions.

\medskip

{
\small

\bibliography{neurips_2021}
\bibliographystyle{plain}

}

\newpage

\appendix

\input{appendix}

\end{document}

%% file: results_table_single.tex
\begin{table}[t!]
    \small
    \centering
    
    \caption{ 
    Results on four different architectures and three datasets: Cifar-10, Cifar-100 and Tiny ImageNet. Test accuracy (in percentages) for time budgets: 25\%, 50\%, 75\%, 100\% of the base network, and Max without any time limits. 
    The first column shows the test accuracy of the base network. The results represent a mean of three runs and standard deviations are provided in Appendix \ref{sec:all_plots}. We bold results within two standard deviations of the best model.} 
    
    \vspace{1em}
    \begin{subtable}[h]{0.49\textwidth}
    \begin{tabular}{@{}c@{\;}c@{}rrrrr}
         
         \toprule
         \multicolumn{7}{c}{\textbf{ResNet-56}} \\
         
         \midrule
         Data & Algo &  \hspace{0.25em}  25\% & 50\% & 75\% & 100\% & Max \\
         \midrule
         
         \multirow{3}{*}{\begin{tabular}[x]{@{}c@{}}\textbf{C10}\\($92.0 $)\end{tabular}}
         & SDN &
         $77.7 $ & $87.3 $ & $91.1 $ & $\mathbf{92.0} $ & $\mathbf{92.1} $ \\
         
         & PBEE
         & $69.8 $ & $81.8 $ & $87.5 $ & $91.0 $ & $\mathbf{92.1} $
         \\
         
        & \ourabb{}
        & $\mathbf{80.3} $ & $\mathbf{88.7}$ & $\mathbf{91.5} $ & $\mathbf{92.1} $ & $\mathbf{92.1} $
        \\

         \midrule
         
         \multirow{3}{*}{\begin{tabular}[x]{@{}c@{}}\textbf{C100}\\($68.4 $)\end{tabular}}
          & SDN
          & $47.1 $ & $57.2 $ & $64.7 $ & $69.0 $ & $69.7 $
          \\
          
          & PBEE
          & $45.2 $ & $53.5 $ & $60.1 $ & $67.0 $ & $69.0 $
          \\
          
          & \ourabb{}
          & $\mathbf{51.3} $ & $\mathbf{62.1} $ & $\mathbf{68.4} $ & $\mathbf{70.7} $ & $\mathbf{70.9} $
          \\
         \midrule 
         
         \multirow{3}{*}{\begin{tabular}[x]{@{}c@{}}\textbf{T-IM}\\($53.9 $)\end{tabular}}
         & SDN
         & $31.2 $ & $41.2 $ & $49.9 $ & $54.5 $ & $54.7 $
         \\
         
         & PBEE
         & $29.0 $ & $37.6 $ & $48.2 $ & $53.4 $ & $54.3 $
         \\
         & \ourabb{}
         & $\mathbf{35.2} $ & $\mathbf{46.2} $ & $\mathbf{53.7} $ & $ \mathbf{56.3} $ & $\mathbf{56.4} $
         \\ 
         \bottomrule
    \end{tabular}
    \end{subtable}
    \hfill
    \begin{subtable}{0.49\textwidth}
    \begin{tabular}{@{}c@{\;}c@{}rrrrr}
        \toprule
         \multicolumn{7}{c}{\textbf{MobileNet}} \\
         \midrule
         Data & Algo & \hspace{0.25em}  25\% & 50\% & 75\% & 100\% & Max \\
         \midrule
         
         \multirow{3}{*}{\begin{tabular}[x]{@{}c@{}}\textbf{C10}\\($90.6 $)\end{tabular}}
         & SDN 
         & $\mathbf{86.1} $ & $\mathbf{90.5} $ & $90.8 $ & $90.7 $ & $90.9 $
         \\
         
         & PBEE 
         & $76.3 $ & $85.9 $ & $89.7 $ & $90.9 $ & $91.1 $
         \\
         
         & \ourabb{} 
         & $\mathbf{86.7} $ & $\mathbf{90.9} $ & $\mathbf{91.4} $ & $\mathbf{91.4} $ & $\mathbf{91.5} $
         \\
         
         \midrule
         
        \multirow{3}{*}{\begin{tabular}[x]{@{}c@{}}\textbf{C100}\\($65.1 $)\end{tabular}}
        & SDN 
        & $\mathbf{54.3} $ & $63.5 $ & $66.8 $ & $67.8 $ & $67.9 $
        \\
        
        & PBEE 
        & $47.1 $ & $61.6 $ & $61.6 $ & $67.0 $ & $68.0 $
        \\
        
        & \ourabb{} 
        & $\mathbf{54.5} $ & $\mathbf{65.2} $ & $\mathbf{68.4} $ & $\mathbf{69.0} $ & $\mathbf{69.1} $
        \\
        
         \midrule 
         
        \multirow{3}{*}{\begin{tabular}[x]{@{}c@{}}\textbf{T-IM}\\($59.3 $)\end{tabular}}
        & SDN 
        & $\mathbf{35.6} $ & $\mathbf{47.1} $ & $55.3 $ & $58.9 $ & $59.7 $
        \\
        
        & PBEE 
        & $26.7 $ & $38.4 $ & $50.3 $ & $55.6 $ & $59.7 $
        \\
        
        & \ourabb{} 
        & $\mathbf{37.3} $ & $\mathbf{49.5} $ & $\mathbf{56.7} $ & $\mathbf{59.7} $ & $\mathbf{60.2} $
        \\

         \bottomrule
    \end{tabular}
    \end{subtable}

    \begin{subtable}{0.49\textwidth}
    \begin{tabular}{@{}c@{\;}c@{}rrrrr}
            \toprule
             \multicolumn{7}{c}{\textbf{WideResNet}} \\
         \midrule
         Data & Algo &  \hspace{0.25em}   25\% & 50\% & 75\% & 100\% & Max \\
         \midrule

         \multirow{3}{*}{\begin{tabular}[x]{@{}c@{}}\textbf{C10}\\($94.4 $)\end{tabular}}
         & SDN 
         & $83.8 $ & $91.7 $ & $94.1 $ & $94.4 $ & $94.4 $
         \\
         
         & PBEE
         & $78.0 $ & $84.0 $ & $90.3 $ & $93.8 $ & $94.4 $
         \\
         
         & \ourabb{} 
         & $\mathbf{86.7} $ & $\mathbf{92.9} $ & $\mathbf{94.5} $ & $\mathbf{94.7} $ & $\mathbf{94.7} $
         \\
         
         \midrule
         
         \multirow{3}{*}{\begin{tabular}[x]{@{}c@{}}\textbf{C100}\\($75.1 $)\end{tabular}}
         & SDN 
         & $55.9 $ & $65.1 $ & $71.6 $ & $75.0 $ & $75.4 $
         \\
         
         & PBEE 
         & $46.7 $ & $57.2 $ & $66.0 $ & $73.2 $ & $75.4 $
         \\
         
         & \ourabb{} 
         & $\mathbf{59.5} $ & $\mathbf{69.1} $ & $\mathbf{74.5} $ & $\mathbf{76.2} $ & $\mathbf{76.4} $
         \\

         \midrule 
         
         \multirow{3}{*}{\begin{tabular}[x]{@{}c@{}}\textbf{T-IM}\\($59.6 $)\end{tabular}}
         & SDN 
         & $36.8 $ & $46.0 $ & $54.6 $ & $59.4 $ & $59.7 $
         \\
         
         & PBEE 
         & $29.9 $ & $37.8 $ & $52.7 $ & $58.5 $ & $59.7 $
         \\
         
         & \ourabb{} 
         & $\mathbf{40.0} $ & $\mathbf{50.1} $ & $\mathbf{57.5} $ & $\mathbf{60.2} $ & $\mathbf{60.3} $
         \\
         
         \bottomrule 
         
    \end{tabular}
    \end{subtable}
    \hfill
    \hspace{0.2em}
    \begin{subtable}{0.49\textwidth}
    \begin{tabular}{@{}c@{\;}c@{}rrrrr}
            \toprule
             \multicolumn{7}{c}{\textbf{VGG}} \\
         \midrule
         Data & Algo & \hspace{0.25em}    25\% & 50\% & 75\% & 100\% & Max \\
         \midrule
         
        \multirow{3}{*}{\begin{tabular}[x]{@{}c@{}}\textbf{C10}\\($93.0 $)\end{tabular}}
        & SDN 
        & $86.0 $ & $92.1 $ & $\mathbf{93.0} $ & $\mathbf{93.0} $ & $\mathbf{93.0} $
        \\
        
        & PBEE
        & $75.0 $ & $86.0 $ & $91.0 $ & $92.9 $ & $\mathbf{93.1} $
        \\
        
        & \ourabb{} 
        & $\mathbf{87.1} $ & $\mathbf{92.5} $ & $\mathbf{93.2} $ & $\mathbf{93.2} $ & $\mathbf{93.2} $
        \\
         \midrule
         
        \multirow{3}{*}{\begin{tabular}[x]{@{}c@{}}\textbf{C100}\\($70.4 $)\end{tabular}}
        & SDN 
        & $58.5 $ & $67.2 $ & $70.6 $ & $71.4 $ & $71.5 $
        \\
        
        & PBEE
        & $51.2 $ & $65.3 $ & $65.3 $ & $70.9 $ & $72.0 $
        \\
        
        & \ourabb{} 
        & $\mathbf{60.2} $ & $\mathbf{69.3} $ & $\mathbf{72.6} $ & $\mathbf{73.5} $ & $\mathbf{73.6} $
        \\
        
         \midrule
         
        \multirow{3}{*}{\begin{tabular}[x]{@{}c@{}}\textbf{T-IM}\\($59.0 $)\end{tabular}}
        & SDN 
        & $40.0 $ & $50.5 $ & $57.4 $ & $\mathbf{59.6} $ & $59.7 $
        \\
        
        & PBEE
        & $31.0 $ & $45.2 $ & $55.2 $ & $\mathbf{60.1} $ & $\mathbf{60.2} $
        \\
        
        & \ourabb{} 
        & $\mathbf{41.4} $ & $\mathbf{52.3} $ & $\mathbf{59.3} $ & $\mathbf{60.1} $ & $\mathbf{60.5} $
        \\

         \bottomrule
    \end{tabular}
    \end{subtable}
    \label{tab:results_table}
\end{table}

%% file: appendix.tex
\section{Training Details}
All experiments were performed using a~single Tesla V100 GPU.

\subsection{Supervised Learning}
\label{sec:training_details}
We setup the core networks in our CIFAR-10, CIFAR-100, and Tiny~ImageNet experiments following~\cite{kaya2019shallow} for fair comparison. We use these trained networks and treat them as pre-trained models, i.e.~we consider the ,,IC-only'' setup, where we do not change the base network.

For CIFAR-10 and CIFAR-100 we train ICs for 50 epochs using the Adam optimizer with learning rate set to $0.001$, but lowered by a~factor of 10 after 15 epochs. When training on Tiny~ImageNet, the learning rate is additionally lowered again by the same factor after epoch 40. On ImageNet (on the pretrained ResNet-50 from the \emph{torchvision} package), the ICs are trained for $40$ epochs, with the initial learning rate of $0.00001$ being reduced by a~factor of 10 in epochs 20 and 30. To train the ensembling part of our method, we run SGD on the training dataset for 500 epochs. Since both the dataset and the model are very small, we use a~high number of epochs to ensure convergence.

\paragraph{Architecture and Placement of ICs}
\label{sec:appendix_placement_ic}
Most common computer vision architectures, including the ones we use, are divided into blocks (e.g. residual blocks in ResNet). Because some blocks change the dimensionality of the features, we take the natural choice of attaching an $\IC$ after each block, which also considerably simplifies the implementation of our method for any future architectures. 
Note that the resulting uniform distribution of $\IC$s along the base network is not necessarily optimal~\cite{scardapane2020should}. However, we focus on this setup for the sake of a~fair comparison with SDN and PBEE and consider the exploration of the best placement of $\IC$s as outside the scope of this work.

Each $\IC$ consists of a~single convolutional layer, a~pooling layer, and a~fully-connected layer, which outputs the class logits. The convolutional layer has a~kernel size of 3 with the number of output filters equal to the number of input channels. When applying \stck{cascade connections} in \our, we use the outputs of the previous $\IC$ as an additional input to the linear classification layer of the current $\IC$, as shown earlier in Figure~\ref{fig:re_intro}. 
Because Tiny ImageNet has a~larger input image size than CIFAR datasets, we use convolutions with stride $2$ instead of $1$ to reduce the number of operations of each $\IC$. 

For the pooling layer we reuse the SDN pooling proposed by~\cite{kaya2019shallow}, which is defined as: 
\begin{equation*}
\mathrm{sdn\_pool}(x) = \gamma \cdot \mathrm{avg\_pool}(x) + (1 - \gamma) \cdot \mathrm{max\_pool}(x),
\end{equation*}
where $\gamma$ is a~learnable scalar parameter. It reduces the size of convolutional maps to $4 \times 4$. 

We keep the architecture and IC placement  fixed between experiments, but with small exceptions for Tiny~ImageNet and ImageNet. For Tiny~ImageNet, we use convolutional layers with stride set to $2$ if all dimensions of the input are larger than 8. We do the same for ImageNet, but we additionally reduce the number of output channels of that convolution by a~factor of $4$ and we place ICs only every third ResNet block. Finally, we apply Layer Normalization to the output of the preceding IC before using it in the final linear layer.


\subsection{Reinforcement Learning}
\label{sec:training_details_rl}
We set the Atari environments as follows. Every fourth frame (frame skipping) and the one immediately before it are max-pooled. The resulting frame is then rescaled to size $84$x$84$ and converted into grayscale. At every step the agent has a~$0.1$ probability of taking the previous action irrespective of the policy probabilities (sticky actions). This is added to introduce stochasticity into the environment to avoid cases when the policy converges to a~simple strategy that results in the same actions taken in every run. Furthermore, the environment termination flag is set when a~life is lost. Finally, the signum function of the reward is taken (reward clipping). The above setup is fairly common and we base our code on the popular Stable Baselines repository~\cite{stable-baselines3}.

Using that environment setup we use the PPO algorithm to train the policy, and then extract the base network by discarding the value network. We use the following PPO hyperparameters: learning rate $2.5 \cdot 10^{-4}$, $128$ steps to run for each environment per update, batch size $256$, $4$ epochs of surrogate loss optimization, clip range ($\epsilon$) $0.1$, entropy coefficient $0.01$, value function coefficient $0.5$, discount factor $0.99$, $0.95$ as the trade-off of bias vs variance factor for Generalized Advantage Estimator~\cite{schulman2015high}, and the maximum value for the gradient clipping $0.5$. The policy is trained for $10^7$ environment time steps in total.

We use the standard 'NatureCNN'~\cite{mnih2015human} architecture with three convolutional layers and a~single fully connected layer. We attach two ICs after the first and the second layer. Similarly as in the supervised setting, each IC has a~single convolutional layer, an~SDN pooling layer and a~fully connected layer. The convolutional layer has stride set to $4$ and preserves the number of channels. 

To train the ICs, the early-exit policy interacts with the environment. In each step, an IC is chosen uniformly, and the action chosen by that IC is taken. However, the $(o, a_p)$ tuple is actually saved to the replay buffer, with $o$ and $a_p$ being the observation and the action of the original policy, respectively. After $128$ concurrent steps on $8$ environments that buffer is used to train the ICs with behavioral cloning. That is, Kullback–Leibler divergence between the PPO policy actions and the IC actions is used as the cost function. This is done for $5$ epochs with batch size set to $64$ and $128$ for cascading stage and geometric ensembling stage, respectively. The entire process is repeated until $10^6$ or more steps in total are taken.

\section{Additional results}
\label{sec:all_plots}
This section contains experimental results which were omitted in the main part of the paper due to page limitations.

\subsection{Supervised Learning}
\label{sec:additional_results_supervised}
For brevity, in the main part of the paper we have only shown a~table summarizing the results of acceleration on multiple architectures and dataset. Here, we provide a~fuller representation of these results. Figures~\ref{fig:supp_cifar10},~\ref{fig:supp_cifar100} and~\ref{fig:supp_tinyimagenet} (at the end of the Appendix) show results of the tested methods on CIFAR-10, CIFAR-100 and Tiny~ImageNet, respectively. Each figure contains plots for the four considered architectures: ResNet-56, MobileNet, WideResNet and VGG16. Plots show that \ourabb{} outperforms SDN and PBEE in almost all settings, which is consistent with the results summarized earlier. Additionally, in Table~\ref{tab:results_table_appendix} we provide summary of the results with standard deviations. Figures~\ref{fig:supp_hindsight_cifar10},~\ref{fig:supp_hindsight_cifar100},~\ref{fig:supp_hindsight_tinyimagenet} show values of Hindsight Improvability for CIFAR-10, CIFAR-100 and Tiny~ImageNet, respectively.

\input{results_table_appendix}

\subsection{Results of ImageNet experiments}
\label{sec:imagenet_results}
\begin{figure}[tb]
    \centering
    \includegraphics[width=0.8\linewidth]{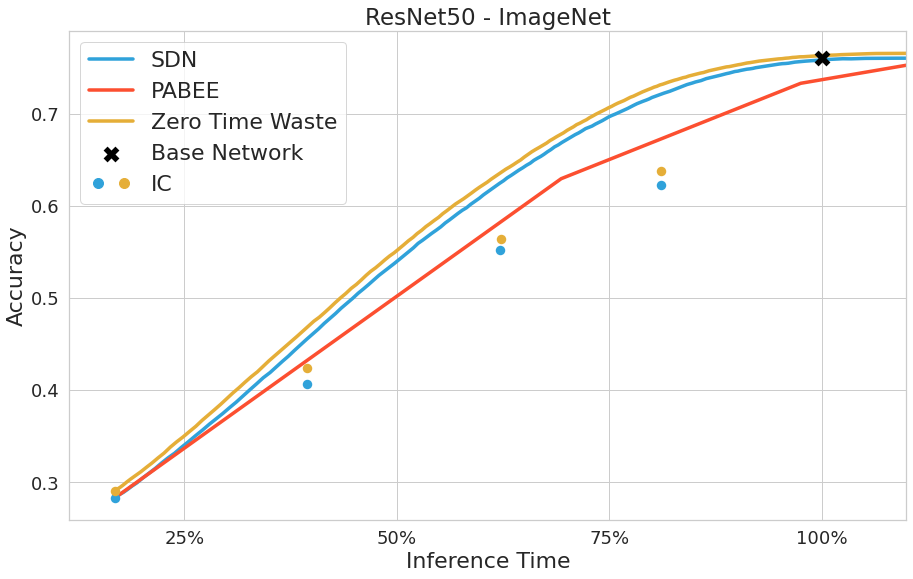}
    \caption{Inference time vs. accuracy for ResNet-50 trained on ImageNet. Base network achieves $76.0\%$ accuracy, and given the same inference time constraint SDN obtains $75.8\%$, PBEE $73.3\%$, and \ourabb{} $76.3\%$.}
    \label{fig:imagenet}
\end{figure}

In order to show that the proposed method scales up well to the ImageNet dataset, we use our method on a~pre-trained model provided by the torchvision package\footnote{\url{https://pytorch.org/vision/stable/index.html}}. The obtained model allows for significant speed-ups on ImageNet while maintaining the same accuracy for the original inference time limit. The results presented in Figure~\ref{fig:imagenet} show that \ourabb{} again outperforms the rest of the methods, with SDN maintaining reasonable, although lower, performance and PBEE generally failing. 
We want to highlight the fact that the architecture of $\IC$s used here is very simple and nowhere as intensely investigated as the architecture of ResNet or other common deep learning models. Adjusting the $\IC$s for this problem could thus improve the results significantly, although we consider this outside the scope of this work. 

\subsection{Results of Transfer Learning experiments}
We investigate whether early exit methods work in a transfer learning setting. We use ResNet-50 from the torchvision package pre-trained on ImageNet similarly as in the previous experiment. To obtain a baseline standard classifier, we remove the final linear layer of the pretrained classifier and train a new linear layer with the number of outputs corresponding to the number of classes in the target dataset. Only then we proceed to train the ICs.


\begin{table}[ht!]
    \small
    \centering
    \caption{ 
    Results on the OCT2017 dataset when using an ImageNet pretrained core network. Test accuracy (in percentages) obtained using the time budget: 25\%, 50\%, 75\%, 100\% of the base network and Max without any limits.}
    \begin{tabular}{@{}c@{\;}ccccc}
        \toprule
        \multicolumn{6}{c}{\textbf{ResNet-50 ($94.6$)}} \\
        \midrule
        Algo &  25\% & 50\% & 75\% & 100\% & Max \\
        \midrule
        SDN & $81.5$ & $93.8$ & $94.6$ & $94.6$ & $94.6$ \\
        PBEE & $56.5$ & $90.3$ & $90.3$ & $94.5$ & $95.2$ \\
        ZTW & $89.4$ & $98.0$ & $98.4$ & $98.5$ & $98.5$ \\
         \bottomrule
    \end{tabular}
    \label{tab:oct_results_table}
\end{table}

We use the OCT-2017 medical dataset \cite{kermany2018identifying} as the target dataset. The training dataset consists of $83484$ high-resolution retinal optical coherence tomography images categorized into four classes, with one class meaning healthy sample, and three diseases. Table \ref{tab:oct_results_table} shows that \ourabb{} outperforms other methods by a significant margin, and manages to cut down the time required to obtain the accuracy of the baseline by over 75\%. This suggests that leveraging the power of previous ICs is especially useful when the features are not perfectly adjusted to the problem at hand, i.e. were trained for ImageNet classification and used for pathology classification data from a completely different domain. We aim to explore the transfer learning setting in future work.

\subsection{Results of Reinforcement Learning experiments}

\begin{figure*}
    \centering
    \begin{subfigure}[t]{0.48\textwidth}
        \includegraphics[width=\textwidth]{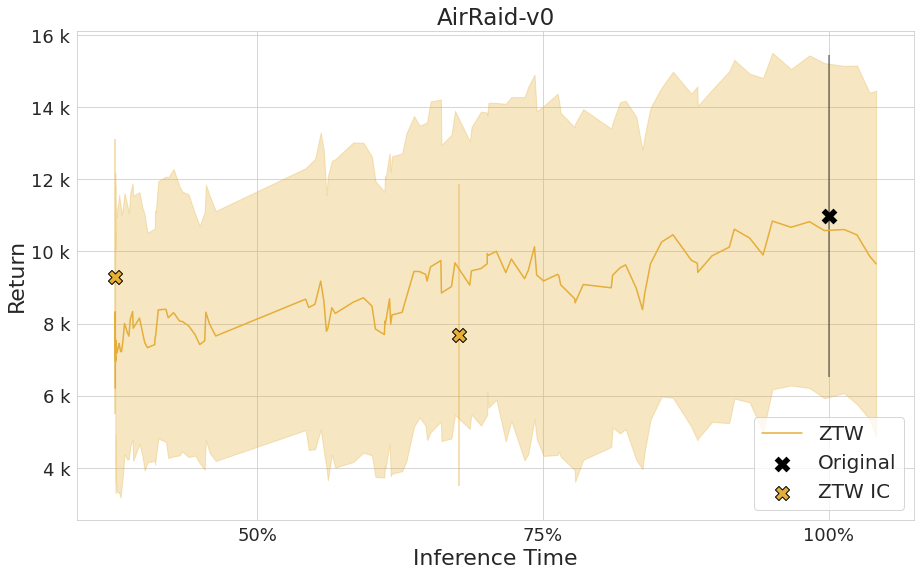}
    \end{subfigure}
    \begin{subfigure}[t]{0.48\textwidth}
        \includegraphics[width=\textwidth]{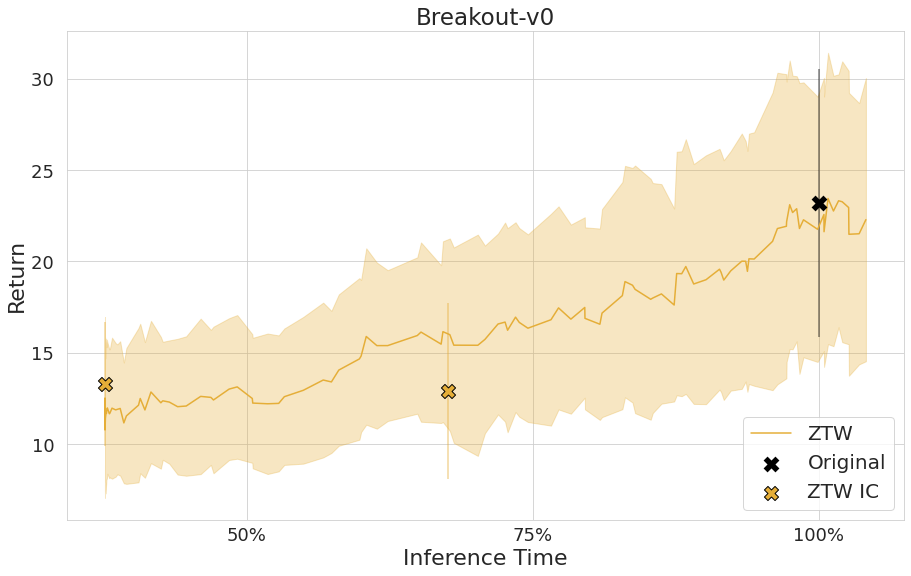}
    \end{subfigure}
    \begin{subfigure}[t]{0.48\textwidth}
        \includegraphics[width=\textwidth]{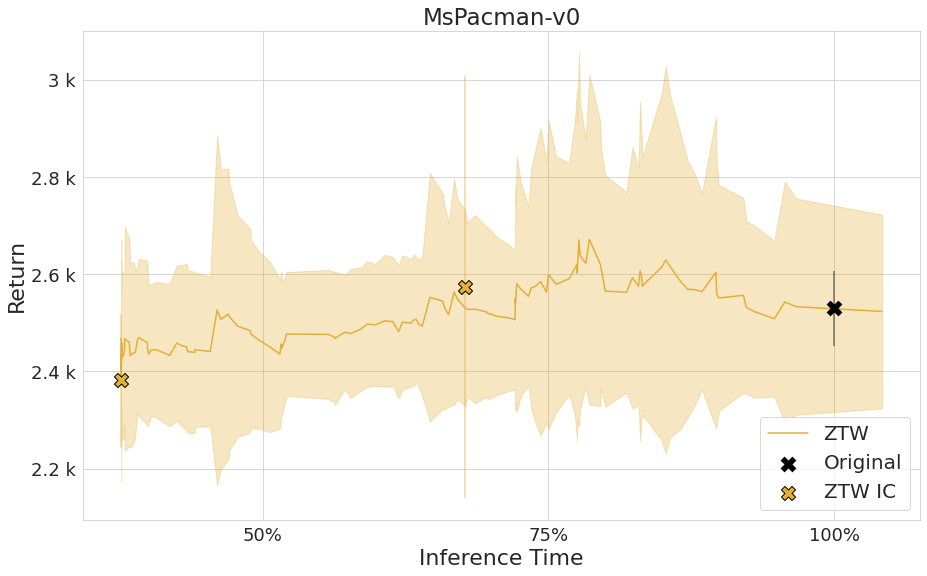}
    \end{subfigure}
    \begin{subfigure}[t]{0.48\textwidth}
        \includegraphics[width=\textwidth]{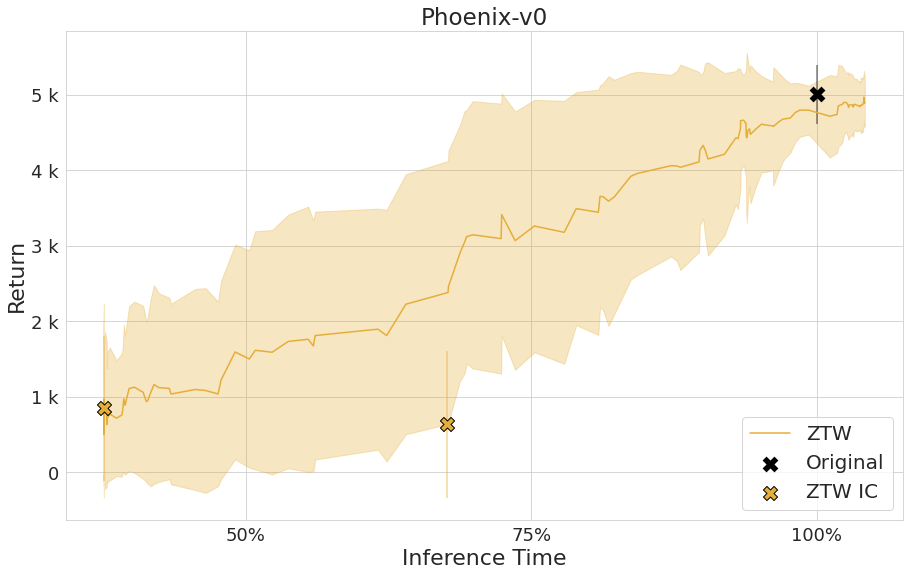}
    \end{subfigure}
    \begin{subfigure}[t]{0.48\textwidth}
        \includegraphics[width=\textwidth]{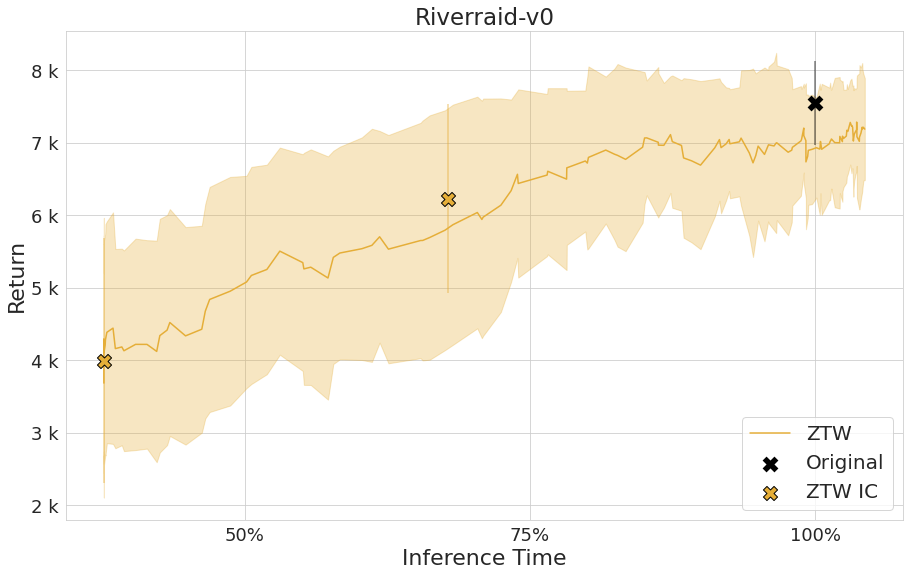}
    \end{subfigure}
    \begin{subfigure}[t]{0.48\textwidth}
        \includegraphics[width=\textwidth]{figs/rl_plots/pong.png}
    \end{subfigure}
    \begin{subfigure}[t]{0.48\textwidth}
        \includegraphics[width=\textwidth]{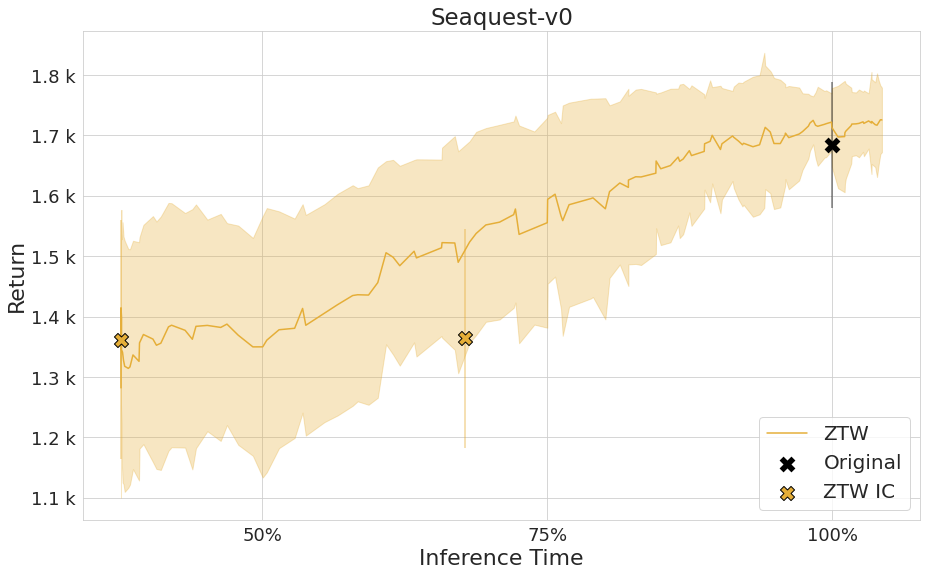}
    \end{subfigure}
    \begin{subfigure}[t]{0.48\textwidth}
        \includegraphics[width=\textwidth]{figs/rl_plots/qbert.png}
    \end{subfigure}
    
    \caption{Mean and standard deviation of returns for multiple confidence thresholds on various Atari 2600 environments. Some environments allow significant computational savings with a~negligible or no impact on performance.}
    \label{fig:rl_rest_results}
\end{figure*}

In Figure~\ref{fig:rl_rest_results} we show the results for all eight Reinforcement Learning environments that we ran our experiments on. Degree of time savings depends heavily on the environment. For some of the environments, such as AirRaid and Pong, the ICs obtain a~similar return to that of the original policy. Because of that the resulting plot is almost flat, allowing for significant inference time reduction without any performance drop. Other environments, such as Seaquest, Phoenix and Riverraid, allow to gradually trade-off performance for inference time just as in the supervised setting.

\section{Ablation Studies}

In this section, we present results of experiments which explain our design decisions. In particular, we focus here on four issues: (1) what is the individual impact of cascade connections and geometric ensembling, (2) how performance of additive and geometric ensembles compares in our setting, (3) how stopping the gradient in \stck{cascade connections} impacts learning dynamics, and (4) how the number of classes in the training dataset impacts the results.
\label{sec:ablation}

\subsection{Impact of cascading and ensembling}

\begin{figure}[t]
    \centering
    \includegraphics[width=0.9\linewidth]{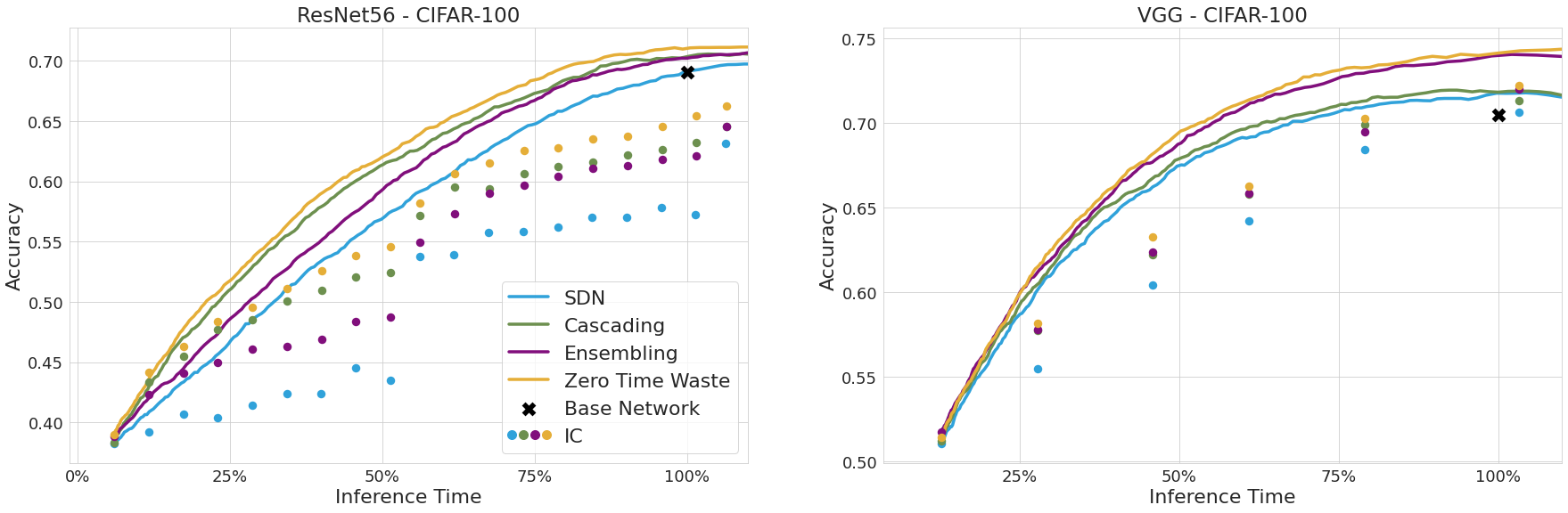}
    \caption{Ablation studies exhibiting the importance of both techniques proposed in the paper. Although both \stck{cascade connections} and geometric ensembling seem to help, the exact effect depends on the architecture and chosen threshold $\tau$. For ResNet56 \stck{cascade connections} seem to be much more helpful than ensembling, while for VGG16 the opposite is true. As such, both are required to consistently improve results.}
    \label{fig:ablation}
\end{figure}

An important question is whether we need both components in the proposed model (\stck{cascade connections} and ensembling), and what role do they play in the final performance of our model. Figure~\ref{fig:ablation} shows the results of independently applied \stck{cascade connections} and geometric ensembling on a~ResNet-56 and VGG-16 trained on CIFAR-100. We observe that depending on the threshold $\tau$ and the architecture, one of these techniques may be more important than the other. However, combining these methods consistently improves the performance each of them achieved independently. Thus we argue that both \stck{cascade connections} and geometric ensembling are required in \our{} and using only one of them will lead to significant performance deterioration.

\subsection{Geometric vs Additive Ensembles}
\label{sec:additive_vs_geometric}

In this work we proposed geometric ensembles for combining predictions from multiple $\IC$s. Here, we show how this approach performs in comparison to additive ensemble of the form:

\begin{equation}
\label{eq:appendix_ensemble_eq}
    q_m^i(x){}=\frac{1}{Z_m}\,{}\sum_{j\leq{}m}w_m^j p_j^i(x) + b_m^i,
\end{equation}
where $b_m^i > 0$ and $w_m^j>0$, for $j=1,\dots,m$, are trainable parameters, and $Z_m$ is a~normalization value, such that $\sum_i q_m^i(x) = 1$. That is, we use the same approach as in geometric ensembles, but we substitute the product for a~sum and change the weighting scheme.

\begin{figure*}
    \centering
    \begin{subfigure}[t]{0.48\textwidth}
        \includegraphics[width=\textwidth]{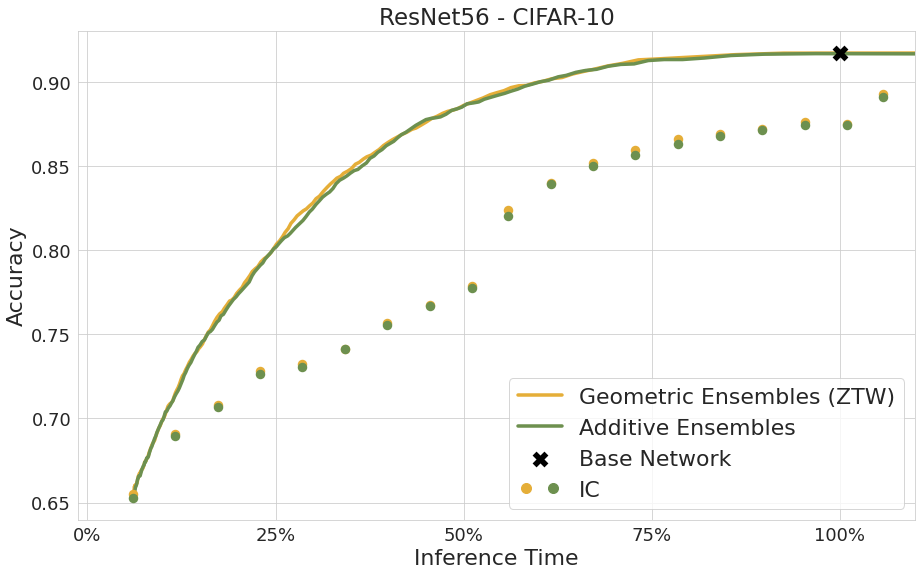}
    \end{subfigure}
    \begin{subfigure}[t]{0.48\textwidth}
        \includegraphics[width=\textwidth]{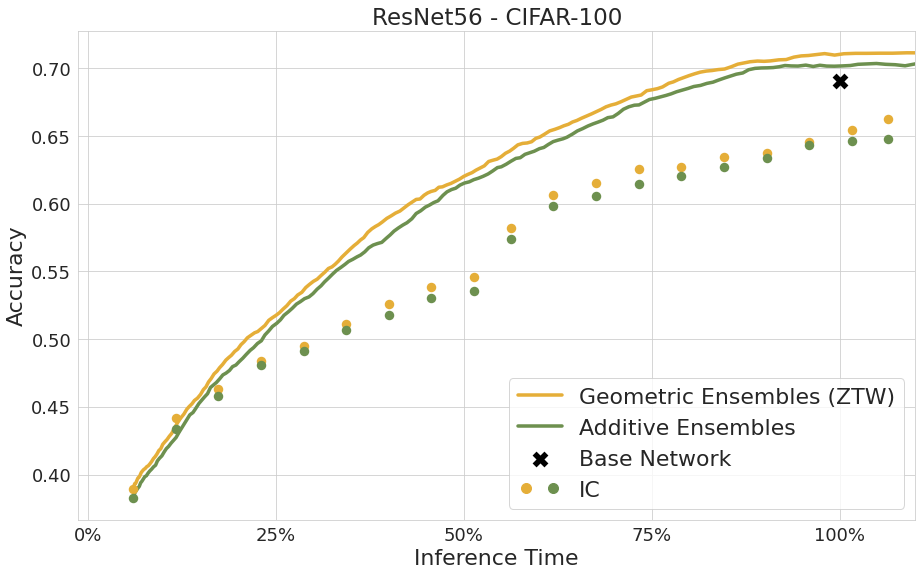}
    \end{subfigure}
    
    \begin{subfigure}[t]{0.48\textwidth}
        \includegraphics[width=\textwidth]{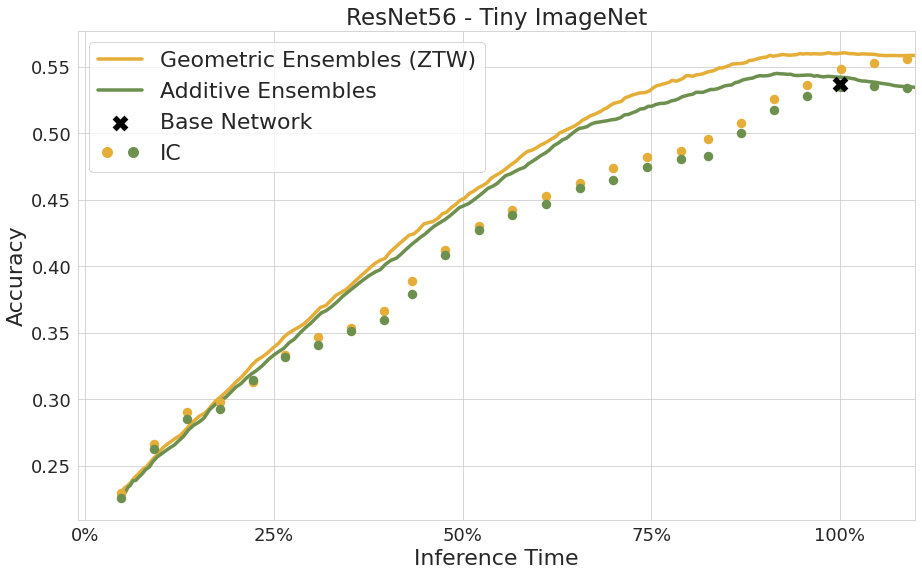}
    \end{subfigure}

    \caption{Comparison of geometric and additive ensembling on ResNet-56 with \stck{cascade connections}, conducted on CIFAR-10, CIFAR-100, and Tiny~ImageNet.}
    \label{fig:supp_ensembletype}
\end{figure*}

The empirical comparison between an additive ensemble and a~geometric ensemble on ResNet-56 is presented in Figure~\ref{fig:supp_ensembletype}. The results show that the geometric ensemble consistently outperforms the additive ensemble, although the magnitude of improvement varies across datasets. While the difference on CIFAR-10 is negligible, it becomes evident on Tiny ImageNet, especially with the later layers. The results suggest that geometric ensembling is more helpful on more complex datasets with a~larger number of classes.

\subsection{Stop gradients in \stck{cascade connections}}
\label{sec:stop_gradient}

\begin{figure*}
    \centering
    \begin{subfigure}[t]{0.48\textwidth}
        \includegraphics[width=\textwidth]{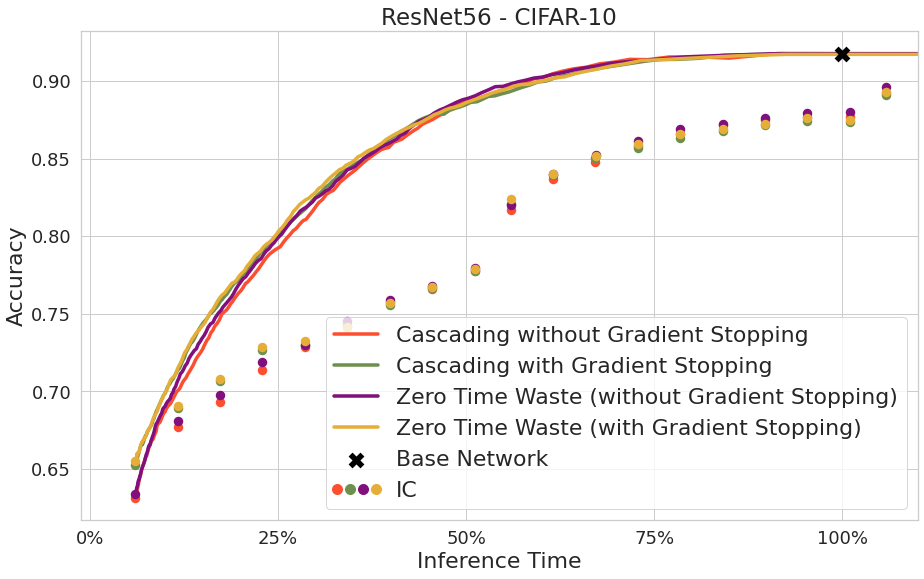}
    \end{subfigure}
    \begin{subfigure}[t]{0.48\textwidth}
        \includegraphics[width=\textwidth]{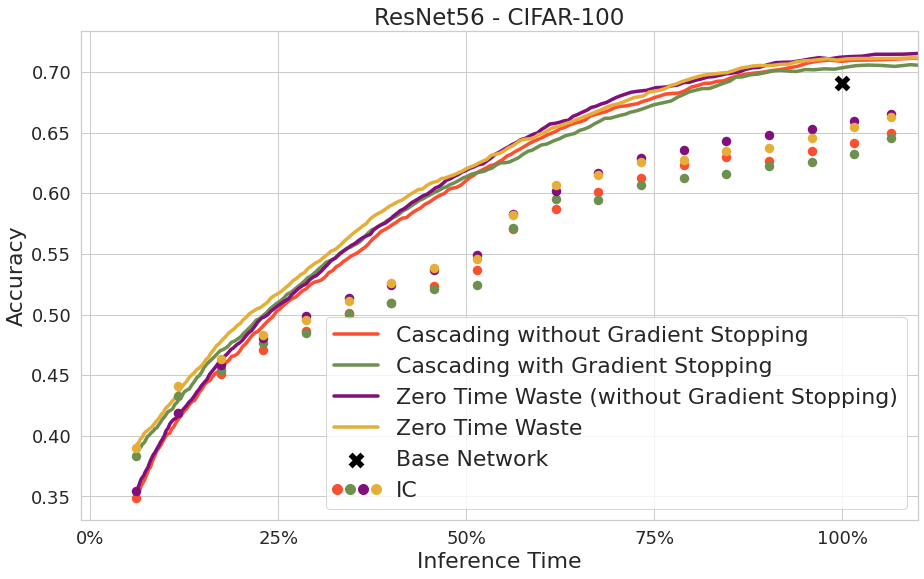}
    \end{subfigure}
    
    \begin{subfigure}[t]{0.48\textwidth}
        \includegraphics[width=\textwidth]{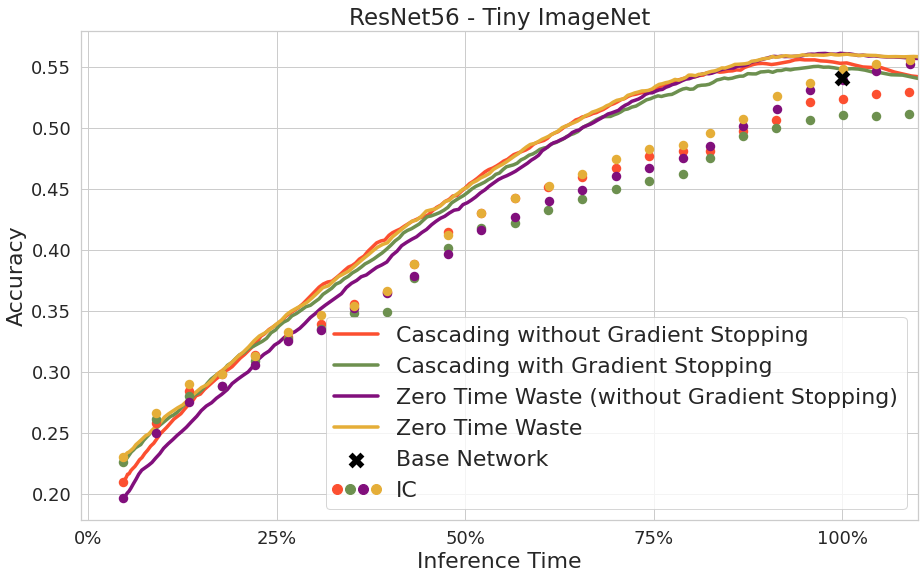}
    \end{subfigure}

    \caption{Effects of stopping gradient in ResNet-56 trained on CIFAR-10, CIFAR-100, and Tiny ImageNet.}
    \label{fig:supp_stopgrad}
\end{figure*}

As mentioned in Section 3 of the main paper, we decide to stop gradient from flowing through the \stck{cascade connections}. We motivate this decision by noticing that the gradients of later layers might destroy the predictive power of the earlier layers. In order to test this hypothesis empirically, we run our experiments on ResNet-56, with and without gradient stopping. As shown in Figure~\ref{fig:supp_stopgrad}, the accuracy of the early $\IC$s is lower when not using gradient stopping. Performance of later $\IC$s may vary, as not using stopping gradient allows greater expressivity for later $\IC$s. Since the second component of our method, ensembling, is able to reuse information from the early $\IC$s we find it beneficial to use gradient stopping in the final model. This is especially evident on Tiny~ImageNet, where on later $\IC$s \stck{cascade connections} perform better without gradient stopping, but \ourabb{} is able to reuse $\IC$s trained with gradient stopping more effectively.

We provide a~more in-depth observation of the reason why the gradient of later $\IC$s might have a~detrimental effect on the performance of early $\IC$s. Observe that in the setting without the detach the parameters of the first $\IC$ will be updated using $\sum_k g_k$, where $g_k$ is the gradient of the loss of the $k$-th $\IC$ wrt. parameters of the first $\IC$. Experimental investigation showed that the cosine similarity of $\sum_k g_k$ and $g_1$ is approximately $0.5$ at the beginning of the training, which means that these gradients point in different directions. Since the gradient $g_1$ represents the best direction for improving the first $\IC$, using $\sum_k g_k$ will lead to a~non-optimal update of its weights, thus reducing its predictive performance. With detach, $g_2 = g_3 = \ldots = 0$ and as such the cosine similarity is always~$1$. This reasoning can be extended to the rest of $\IC$s.

\subsection{Impact of the number of classes}

Additionally, we check how the number of classes in the given problem impacts the results of each method.
To do this, we take the CIFAR-10 dataset, which consists of 10 classes and divide the examples into two more general classes, which can be approximately described as modes of transportation (includes airplane, automobile, horse, ship, truck) and animals (bird, cat, deer, dog, frog). Thus, we obtain a dataset for binary classification which we dub CIFAR-2. We train and evaluate the proposed methods on this dataset with different backbones. Results, summed up in Table \ref{tab:number_of_classes_exp}, show that although performance of ZTW is always on par or better than the baselines, the gap in performance is much smaller, with SDN achieving identical performance in some cases. Although, this might be due to the fact that CIFAR-2 is simpler than original CIFAR-10, we note that \our{} is better suited to non-binary classification problems.

\begin{table}[th!]
    \small
    \centering
    
    \caption{ 
    Results on the CIFAR-2 dataset.} 
    
    \begin{tabular}{@{}c@{\;}cccccc}
         
         \toprule
         \multicolumn{7}{c}{\textbf{ResNet-56}} \\
         
         \midrule
         Data & Algo &  25\% & 50\% & 75\% & 100\% & Max \\
         \midrule
         
         \multirow{3}{*}{\begin{tabular}[x]{@{}c@{}}\textbf{ResNet-56}\end{tabular}}
         
        & SDN & $95.5$ & $96.5$ & $96.5$ & $96.5$ & $96.5$ \\ 
        & PBEE & $91.2$ & $94.1$ & $96.3$ & $96.5$ & $96.6$ \\
        & ZTW & $95.7$ & $96.6$ & $96.6$ & $96.6$ & $96.6$ \\

         \midrule
         
         \multirow{3}{*}{\begin{tabular}[x]{@{}c@{}}\textbf{VGG}\end{tabular}}
         & SDN & $96.6$ & $97.6$ & $97.6$ & $97.6$ & $97.7$ \\ 
         & PBEE & $91.2$ & $96.4$ & $97.2$ & $97.4$ & $97.6$ \\
         & ZTW & $96.7$ & $97.6$ & $97.7$ & $97.7$ & $97.7$ \\

         \midrule 
         
         \multirow{3}{*}{\begin{tabular}[x]{@{}c@{}}\textbf{WideResNet}\end{tabular}}
         & SDN & $95.2$ & $97.0$ & $97.3$ & $97.3$ & $97.4$ \\
         & PBEE & $89.3$ & $93.0$ & $95.9$ & $97.0$ & $97.4$ \\
         & ZTW & $96.3$ & $97.4$ & $97.6$ & $97.6$ & $97.6$ \\
         
         \midrule 
         
         \multirow{3}{*}{\begin{tabular}[x]{@{}c@{}}\textbf{MobileNet}\end{tabular}}
         & SDN & $95.7$ & $96.4$ & $96.4$ & $96.4$ & $96.4$ \\ 
         & PBEE & $91.9$ & $94.3$ & $96.2$ & $96.4$ & $96.4$ \\
         & ZTW & $96.0$ & $96.4$ & $96.4$ & $96.4$ & $96.4$ \\

         \bottomrule
    \end{tabular}
    \label{tab:number_of_classes_exp}
\end{table}

\input{appendix_allplots}

%% file: results_table_appendix.tex
\begin{table}[t!]
    \small
    \centering
    
    \caption{ 
    Results on four different architectures and three datasets: Cifar-10, Cifar-100 and Tiny ImageNet. Test accuracy (in percentages) obtained using the time budget: 25\%, 50\%, 75\%, 100\% of the base network and Max without any limits. 
    The first column shows the test accuracy of the base network.} 
    
    \begin{tabular}{@{}c@{\;}cccccc}
         
         \toprule
         \multicolumn{7}{c}{\textbf{ResNet-56}} \\
         
         \midrule
         Data & Algo &  25\% & 50\% & 75\% & 100\% & Max \\
         \midrule
         
         \multirow{3}{*}{\begin{tabular}[x]{@{}c@{}}\textbf{CIFAR-10}\\($92.0 \pm 0.2$)\end{tabular}}
         & SDN &
         $77.7 \pm 1.0$ & $87.3 \pm 0.5$ & $91.1 \pm 0.2$ & $\mathbf{92.0 \pm 0.1}$ & $\mathbf{92.1 \pm 0.2}$ \\
         
         & PBEE
         & $69.8 \pm 1.3$ & $81.8 \pm 0.3$ & $87.5 \pm 0.1$ & $91.0 \pm 0.3$ & $\mathbf{92.1 \pm 0.3}$
         \\
         
        & \ourabb{}
        & $\mathbf{80.3 \pm 1.0}$ & $\mathbf{88.7 \pm 0.4}$ & $\mathbf{91.5 \pm 0.2}$ & $\mathbf{92.1 \pm 0.3}$ & $\mathbf{92.1 \pm 0.3}$
        \\

         \midrule
         
         \multirow{3}{*}{\begin{tabular}[x]{@{}c@{}}\textbf{CIFAR-100}\\($68.4 \pm 0.2$)\end{tabular}}
          & SDN
          & $47.1 \pm 0.2$ & $57.2 \pm 0.4$ & $64.7 \pm 0.6$ & $69.0 \pm 0.2$ & $69.7 \pm 0.2$
          \\
          
          & PBEE
          & $45.2 \pm 0.5$ & $53.5 \pm 0.5$ & $60.1 \pm 0.5$ & $67.0 \pm 0.2$ & $69.0 \pm 0.2$
          \\
          
          & \ourabb{}
          & $\mathbf{51.3 \pm 0.4}$ & $\mathbf{62.1 \pm 0.3}$ & $\mathbf{68.4 \pm 0.4}$ & $\mathbf{70.7 \pm 0.1}$ & $\mathbf{70.9 \pm 0.1}$
          \\
         \midrule 
         
         \multirow{3}{*}{\begin{tabular}[x]{@{}c@{}}\textbf{Tiny ImageNet}\\($53.9 \pm 0.3$)\end{tabular}}
         & SDN
         & $31.2 \pm 0.2$ & $41.2 \pm 0.3$ & $49.9 \pm 0.4$ & $54.5 \pm 0.5$ & $54.7 \pm 0.4$
         \\
         
         & PBEE
         & $29.0 \pm 0.6$ & $37.6 \pm 0.3$ & $48.2 \pm 0.4$ & $53.4 \pm 0.6$ & $54.3 \pm 0.4$
         \\
         & \ourabb{}
         & $\mathbf{35.2 \pm 0.7}$ & $\mathbf{46.2 \pm 0.4}$ & $\mathbf{53.7 \pm 0.3}$ & $\mathbf{56.3 \pm 0.3}$ & $\mathbf{56.4 \pm 0.3}$
         \\
  \\ 
         \bottomrule
    \end{tabular}
    \begin{tabular}{@{}c@{\;}cccccc}
            \toprule
             \multicolumn{7}{c}{\textbf{MobileNet}} \\
         \midrule
         Data & Algo & 25\% & 50\% & 75\% & 100\% & Max \\
         \midrule
         
         \multirow{3}{*}{\begin{tabular}[x]{@{}c@{}}\textbf{CIFAR-10}\\($90.6 \pm 0.2$)\end{tabular}}
         & SDN 
         & $\mathbf{86.1 \pm 0.5}$ & $\mathbf{90.5 \pm 0.2}$ & $90.8 \pm 0.1$ & $90.7 \pm 0.2$ & $90.9 \pm 0.1$
         \\
         
         & PBEE 
         & $76.3 \pm 0.9$ & $85.9 \pm 0.3$ & $89.7 \pm 0.3$ & $90.9 \pm 0.2$ & $91.1 \pm 0.1$
         \\
         
         & \ourabb{} 
         & $\mathbf{86.7 \pm 0.7}$ & $\mathbf{90.9 \pm 0.3}$ & $\mathbf{91.4 \pm 0.2}$ & $\mathbf{91.4 \pm 0.1}$ & $\mathbf{91.5 \pm 0.1}$
         \\
         
         \midrule
         
        \multirow{3}{*}{\begin{tabular}[x]{@{}c@{}}\textbf{CIFAR-100}\\($65.1 \pm 0.3$)\end{tabular}}
        & SDN 
        & $\mathbf{54.3} \pm 1.4$ & $63.5 \pm 0.8$ & $66.8 \pm 0.4$ & $67.8 \pm 0.1$ & $67.9 \pm 0.1$
        \\
        
        & PBEE 
        & $47.1 \pm 2.7$ & $61.6 \pm 0.7$ & $61.6 \pm 0.7$ & $67.0 \pm 0.3$ & $68.0 \pm 0.3$
        \\
        
        & \ourabb{} 
        & $\mathbf{54.5 \pm 1.1}$ & $\mathbf{65.2 \pm 0.5}$ & $\mathbf{68.4 \pm 0.3}$ & $\mathbf{69.0 \pm 0.1}$ & $\mathbf{69.1 \pm 0.1}$
        \\
        
         \midrule 
         
        \multirow{3}{*}{\begin{tabular}[x]{@{}c@{}}\textbf{Tiny ImageNet}\\($59.3 \pm 0.1$)\end{tabular}}
        & SDN 
        & $\mathbf{35.6 \pm 1.3}$ & $\mathbf{47.1 \pm 0.6}$ & $55.3 \pm 0.3$ & $58.9 \pm 0.2$ & $59.7 \pm 0.1$
        \\
        
        & PBEE 
        & $26.7 \pm 1.5$ & $38.4 \pm 2.0$ & $50.3 \pm 0.8$ & $55.6 \pm 0.3$ & $59.7 \pm 0.0$
        \\
        
        & \ourabb{} 
        & $\mathbf{37.3 \pm 2.8}$ & $\mathbf{49.5 \pm 1.9}$ & $\mathbf{56.7 \pm 0.6}$ & $\mathbf{59.7 \pm 0.4}$ & $\mathbf{60.2 \pm 0.1}$
        \\

         \bottomrule
    \end{tabular}

    \begin{tabular}{@{}c@{\;}cccccc}
            \toprule
             \multicolumn{7}{c}{\textbf{WideResNet}} \\
         \midrule
         Data & Algo &  25\% & 50\% & 75\% & 100\% & Max \\
         \midrule

         \multirow{3}{*}{\begin{tabular}[x]{@{}c@{}}\textbf{CIFAR-10}\\($94.4 \pm 0.1$)\end{tabular}}
         & SDN 
         & $83.8 \pm 1.3$ & $91.7 \pm 0.5$ & $94.1 \pm 0.2$ & $94.4 \pm 0.1$ & $94.4 \pm 0.2$
         \\
         
         & PBEE
         & $78.0 \pm 1.9$ & $84.0 \pm 1.4$ & $90.3 \pm 0.5$ & $93.8 \pm 0.1$ & $94.4 \pm 0.1$
         \\
         
         & \ourabb{} 
         & $\mathbf{86.7 \pm 0.7}$ & $\mathbf{92.9 \pm 0.3}$ & $\mathbf{94.5 \pm 0.1}$ & $\mathbf{94.7 \pm 0.1}$ & $\mathbf{94.7 \pm 0.1}$
         \\
         
         \midrule
         
         \multirow{3}{*}{\begin{tabular}[x]{@{}c@{}}\textbf{CIFAR-100}\\($75.1 \pm 0.1$)\end{tabular}}
         & SDN 
         & $55.9 \pm 1.5$ & $65.1 \pm 0.9$ & $71.6 \pm 0.4$ & $75.0 \pm 0.1$ & $75.4 \pm 0.1$
         \\
         
         & PBEE 
         & $46.7 \pm 2.0$ & $57.2 \pm 1.3$ & $66.0 \pm 0.6$ & $73.2 \pm 0.2$ & $75.4 \pm 0.2$
         \\
         
         & \ourabb{} 
         & $\mathbf{59.5 \pm 0.6}$ & $\mathbf{69.1 \pm 0.9}$ & $\mathbf{74.5 \pm 0.6}$ & $\mathbf{76.2 \pm 0.3}$ & $\mathbf{76.4 \pm 0.2}$
         \\

         \midrule 
         
         \multirow{3}{*}{\begin{tabular}[x]{@{}c@{}}\textbf{Tiny ImageNet}\\($59.6 \pm 0.6$)\end{tabular}}
         & SDN 
         & $36.8 \pm 0.1$ & $46.0 \pm 1.0$ & $54.6 \pm 0.7$ & $59.4 \pm 0.8$ & $59.7 \pm 0.7$
         \\
         
         & PBEE 
         & $29.9 \pm 0.9$ & $37.8 \pm 0.6$ & $52.7 \pm 0.6$ & $58.5 \pm 0.9$ & $59.7 \pm 0.7$
         \\
         
         & \ourabb{} 
         & $\mathbf{40.0 \pm 0.3}$ & $\mathbf{50.1 \pm 0.2}$ & $\mathbf{57.5 \pm 0.4}$ & $\mathbf{60.2 \pm 0.1}$ & $\mathbf{60.3 \pm 0.2}$
         \\
         
         \bottomrule 
         
    \end{tabular}
    \hspace{0.2em}
    \begin{tabular}{@{}c@{\;}cccccc}
            \toprule
             \multicolumn{7}{c}{\textbf{VGG}} \\
         \midrule
         Data & Algo &  25\% & 50\% & 75\% & 100\% & Max \\
         \midrule
         
        \multirow{3}{*}{\begin{tabular}[x]{@{}c@{}}\textbf{CIFAR-10}\\($93.0 \pm 0.0$)\end{tabular}}
        & SDN 
        & $86.0 \pm 0.3$ & $92.1 \pm 0.1$ & $\mathbf{93.0 \pm 0.0}$ & $\mathbf{93.0 \pm 0.0}$ & $\mathbf{93.0 \pm 0.0}$
        \\
        
        & PBEE
        & $75.0 \pm 0.2$ & $86.0 \pm 0.2$ & $91.0 \pm 0.3$ & $92.9 \pm 0.2$ & $93.1 \pm 0.1$
        \\
        
        & \ourabb{} 
        & $\mathbf{87.1 \pm 0.1}$ & $\mathbf{92.5 \pm 0.1}$ & $\mathbf{93.2 \pm 0.2}$ & $\mathbf{93.2 \pm 0.2}$ & $\mathbf{93.2 \pm 0.2}$
        \\
         \midrule
         
        \multirow{3}{*}{\begin{tabular}[x]{@{}c@{}}\textbf{CIFAR-100}\\($70.4 \pm 0.3$)\end{tabular}}
        & SDN 
        & $58.5 \pm 0.4$ & $67.2 \pm 0.1$ & $70.6 \pm 0.3$ & $71.4 \pm 0.2$ & $71.5 \pm 0.4$
        \\
        
        & PBEE
        & $51.2 \pm 0.2$ & $65.3 \pm 0.3$ & $65.3 \pm 0.3$ & $70.9 \pm 0.5$ & $72.0 \pm 0.4$
        \\
        
        & \ourabb{} 
        & $\mathbf{60.2 \pm 0.2}$ & $\mathbf{69.3 \pm 0.4}$ & $\mathbf{72.6 \pm 0.1}$ & $\mathbf{73.5 \pm 0.3}$ & $\mathbf{73.6 \pm 0.4}$
        \\
        
         \midrule
         
        \multirow{3}{*}{\begin{tabular}[x]{@{}c@{}}\textbf{Tiny ImageNet}\\($59.0 \pm 0.2$)\end{tabular}}
        & SDN 
        & $40.0 \pm 1.0$ & $50.5 \pm 0.2$ & $57.4 \pm 0.5$ & $\mathbf{59.6 \pm 0.3}$ & $59.7 \pm 0.3$
        \\
        
        & PBEE
        & $31.0 \pm 1.6$ & $45.2 \pm 0.6$ & $55.2 \pm 0.3$ & $\mathbf{60.1 \pm 0.5}$ & $\mathbf{60.2 \pm 0.5}$
        \\
        
        & \ourabb{} 
        & $\mathbf{41.4 \pm 0.5}$ & $\mathbf{52.3 \pm 0.4}$ & $\mathbf{59.3 \pm 0.4}$ & $\mathbf{60.1 \pm 0.5}$ & $\mathbf{60.5 \pm 0.4}$
        \\

         \bottomrule
    \end{tabular}
    \label{tab:results_table_appendix}
\end{table}

%% file: appendix_allplots.tex
\begin{figure*}[h!]
    \centering
    \begin{subfigure}[t]{0.48\textwidth}
        \includegraphics[width=\textwidth]{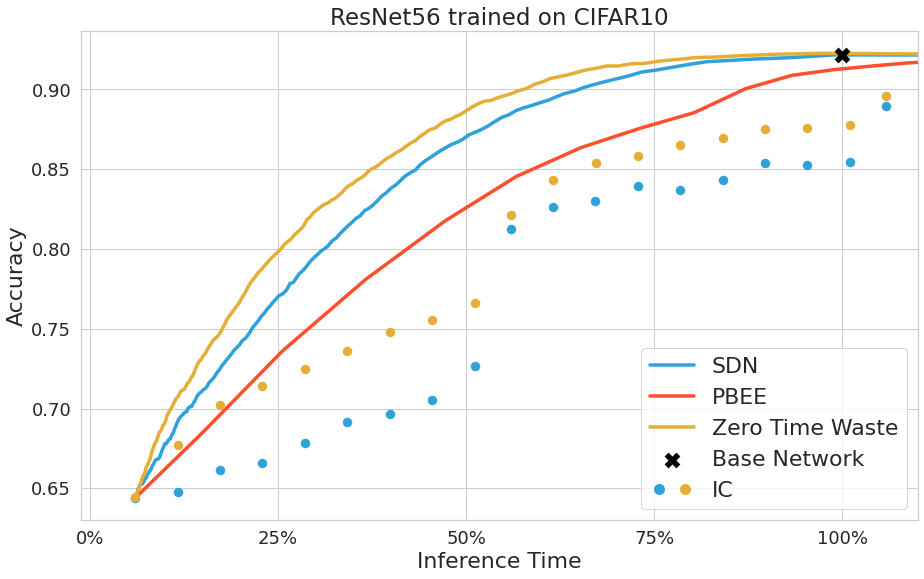}
    \end{subfigure}
    \begin{subfigure}[t]{0.48\textwidth}
        \includegraphics[width=\textwidth]{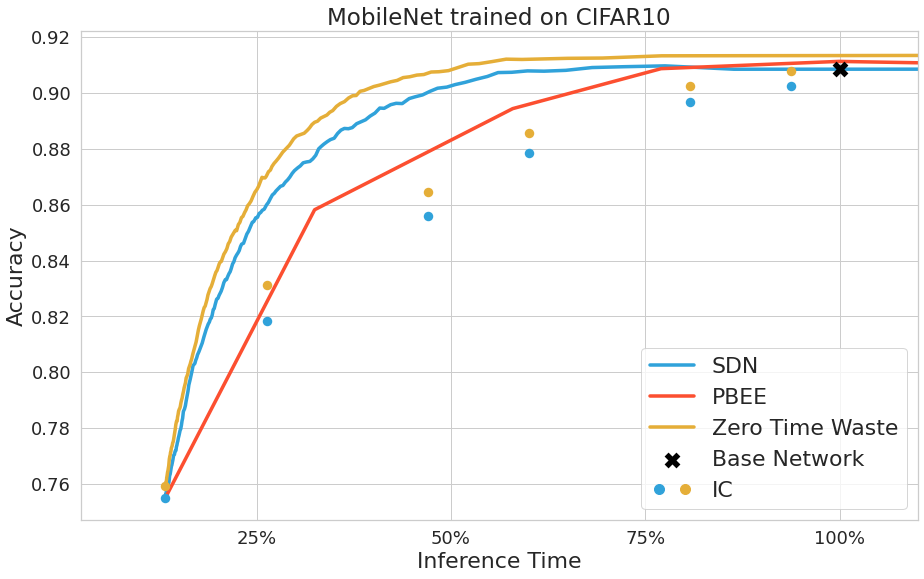}
    \end{subfigure}
    \begin{subfigure}[t]{0.48\textwidth}
        \includegraphics[width=\textwidth]{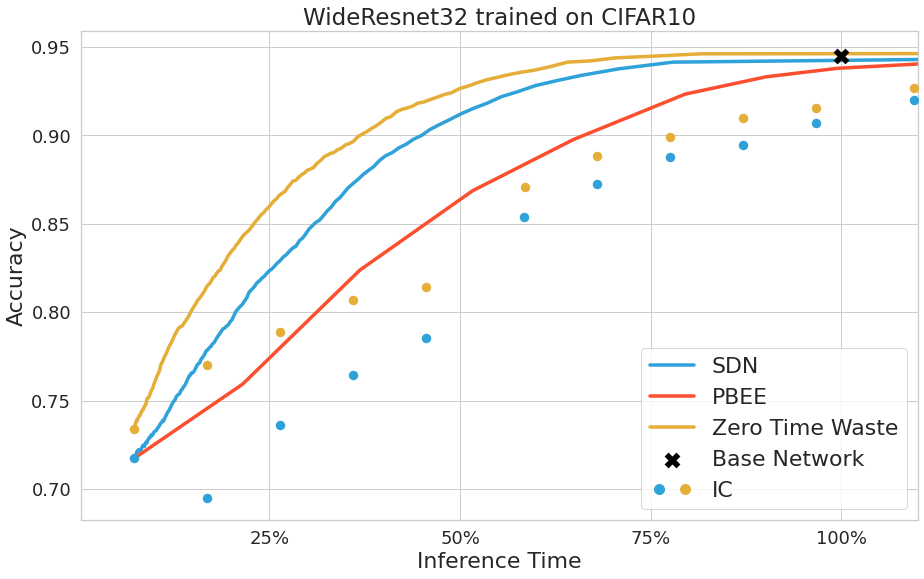}
    \end{subfigure}
    \begin{subfigure}[t]{0.48\textwidth}
        \includegraphics[width=\textwidth]{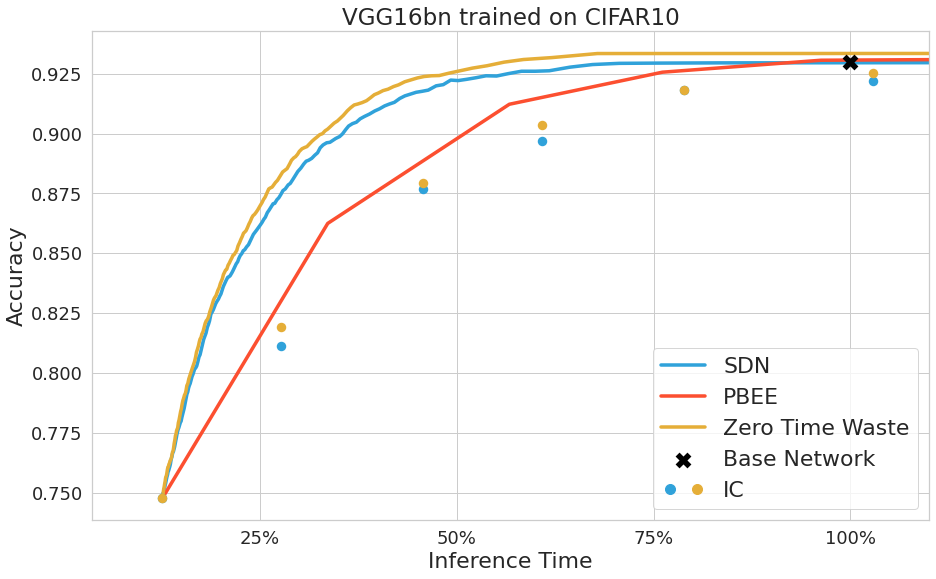}
    \end{subfigure}

    \caption{Inference time vs. accuracy obtained on various architectures trained on CIFAR-10.}
    \label{fig:supp_cifar10}
\end{figure*}

\begin{figure*}[h!]
    \centering
    \begin{subfigure}[t]{0.48\textwidth}
        \includegraphics[width=\textwidth]{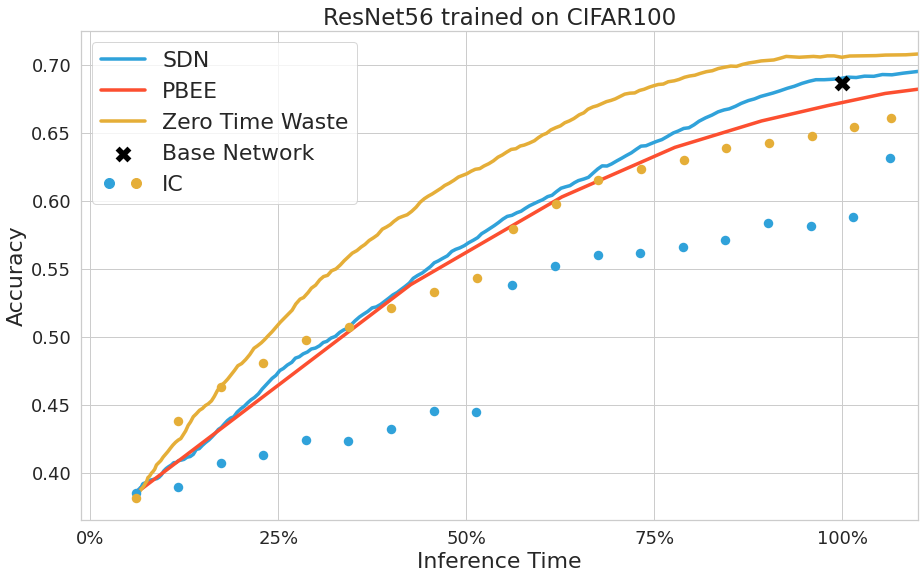}
    \end{subfigure}
    \begin{subfigure}[t]{0.48\textwidth}
        \includegraphics[width=\textwidth]{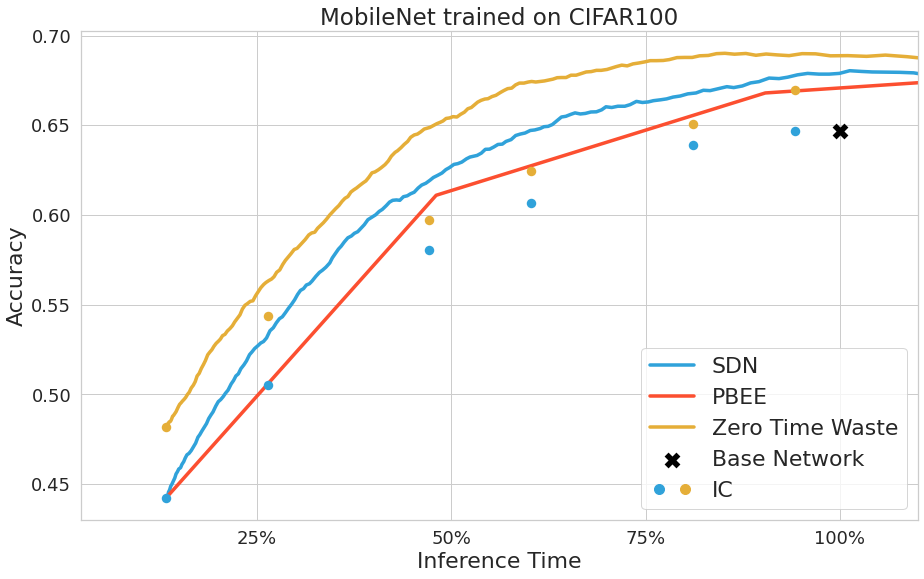}
    \end{subfigure}
    \begin{subfigure}[t]{0.48\textwidth}
        \includegraphics[width=\textwidth]{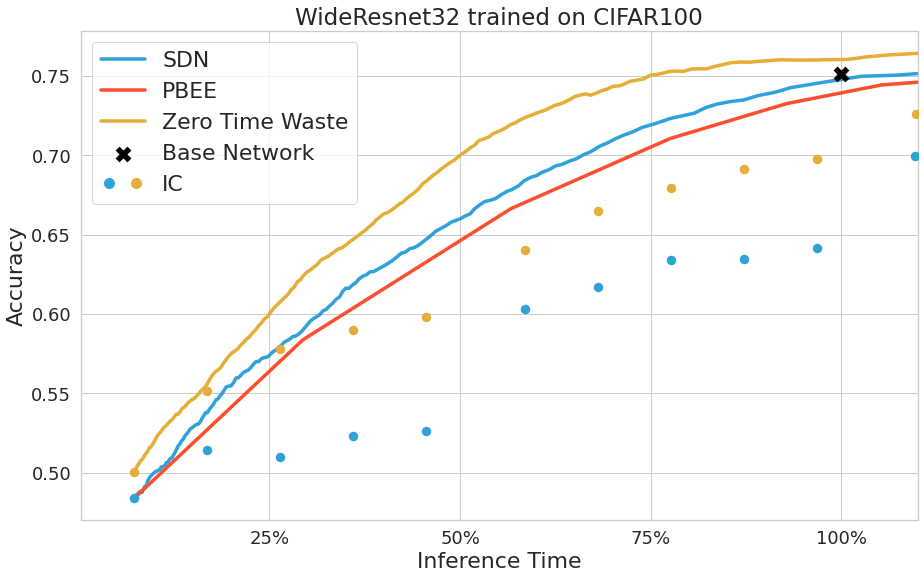}
    \end{subfigure}
    \begin{subfigure}[t]{0.48\textwidth}
        \includegraphics[width=\textwidth]{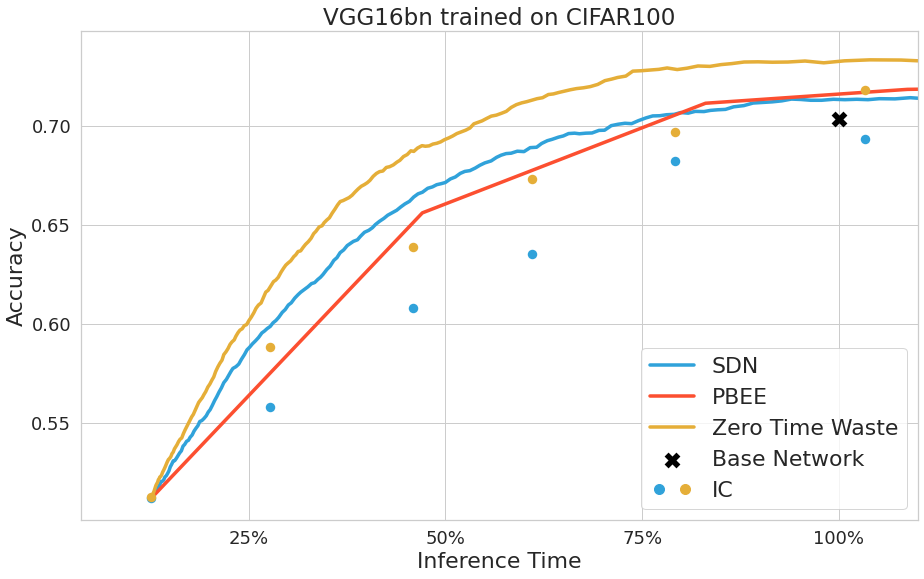}
    \end{subfigure}

    \caption{Inference time vs. accuracy obtained on various architectures trained on CIFAR-100.}
    \label{fig:supp_cifar100}
\end{figure*}

\begin{figure*}[h!]
    \centering
    \begin{subfigure}[t]{0.48\textwidth}
        \includegraphics[width=\textwidth]{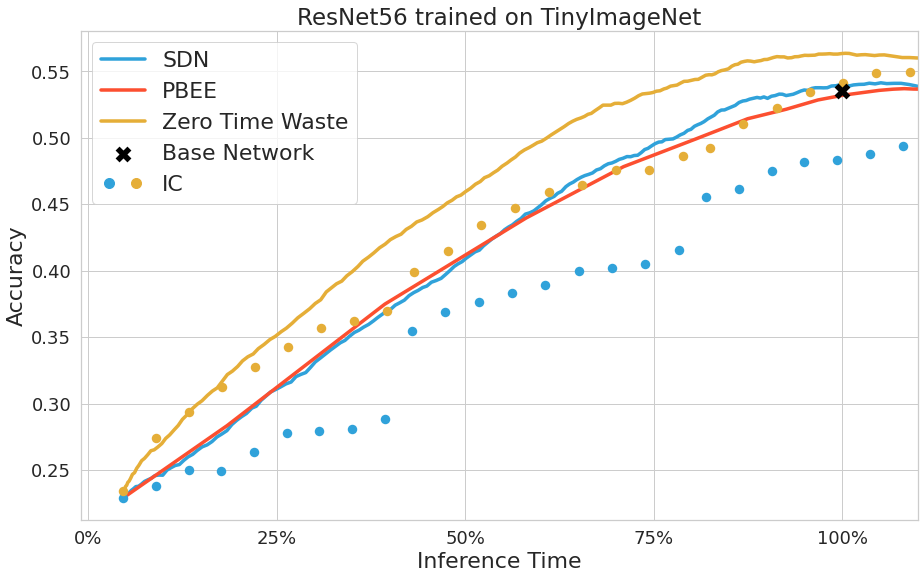}
    \end{subfigure}
    \begin{subfigure}[t]{0.48\textwidth}
        \includegraphics[width=\textwidth]{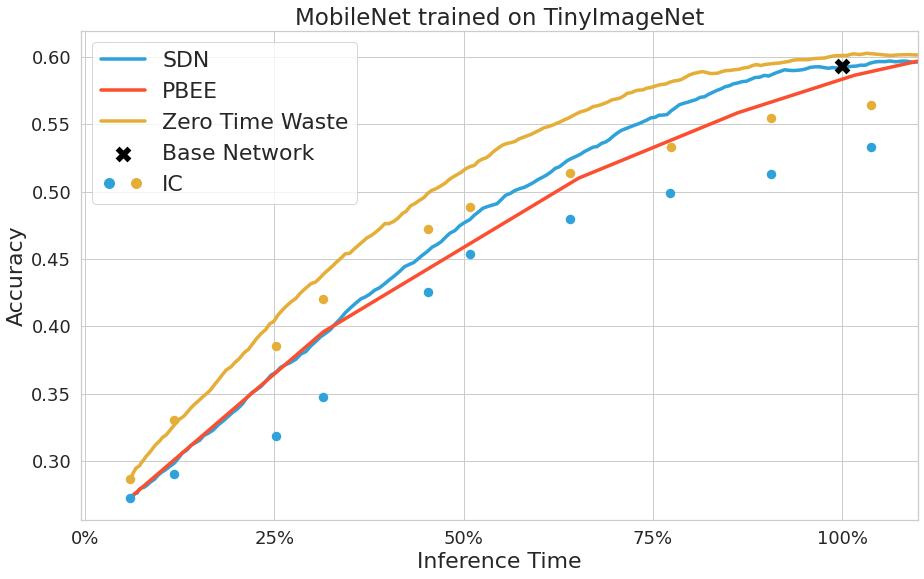}
    \end{subfigure}
    \begin{subfigure}[t]{0.48\textwidth}
        \includegraphics[width=\textwidth]{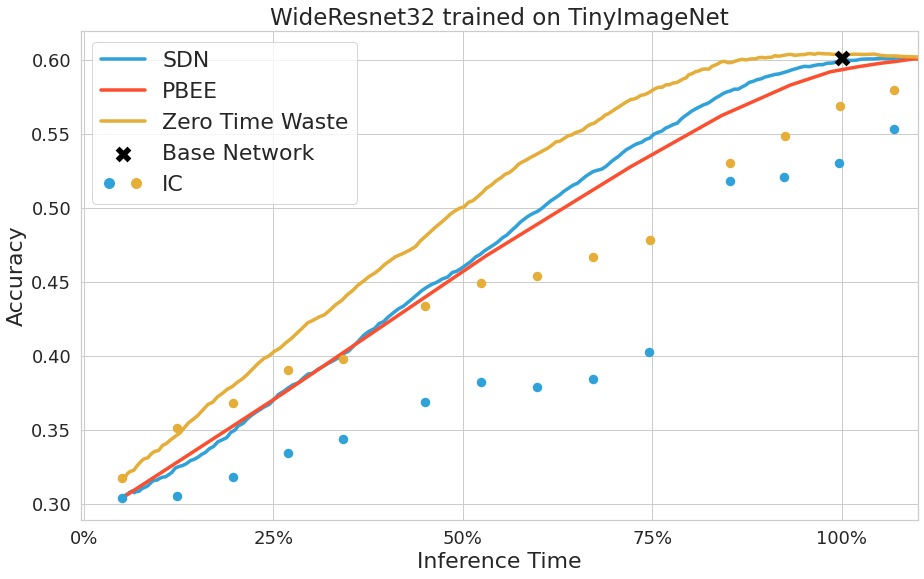}
    \end{subfigure}
    \begin{subfigure}[t]{0.48\textwidth}
        \includegraphics[width=\textwidth]{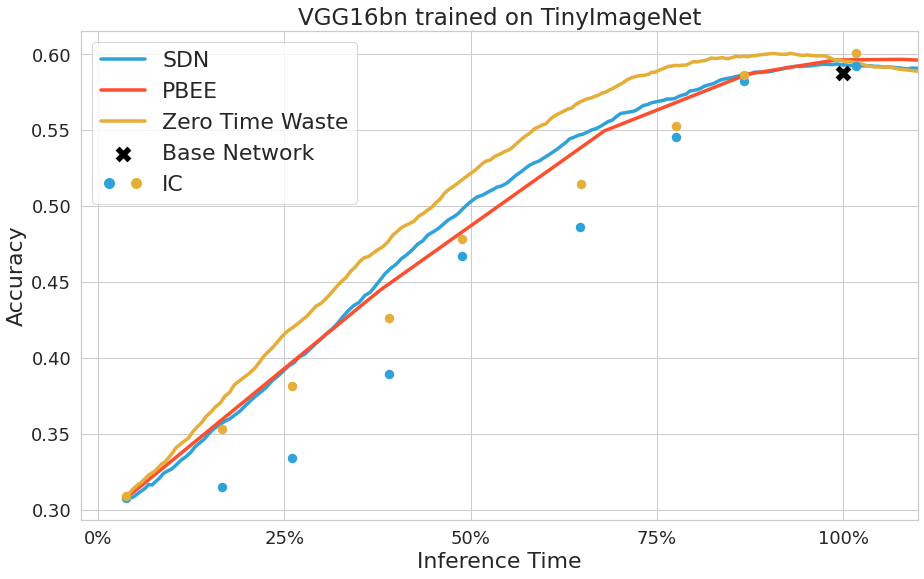}
    \end{subfigure}

    \caption{Inference time vs. accuracy obtained on various architectures trained on Tiny~ImageNet.}
    \label{fig:supp_tinyimagenet}
\end{figure*}


\begin{figure*}[h!]
    \centering
    \begin{subfigure}[t]{0.48\textwidth}
        \includegraphics[width=\textwidth]{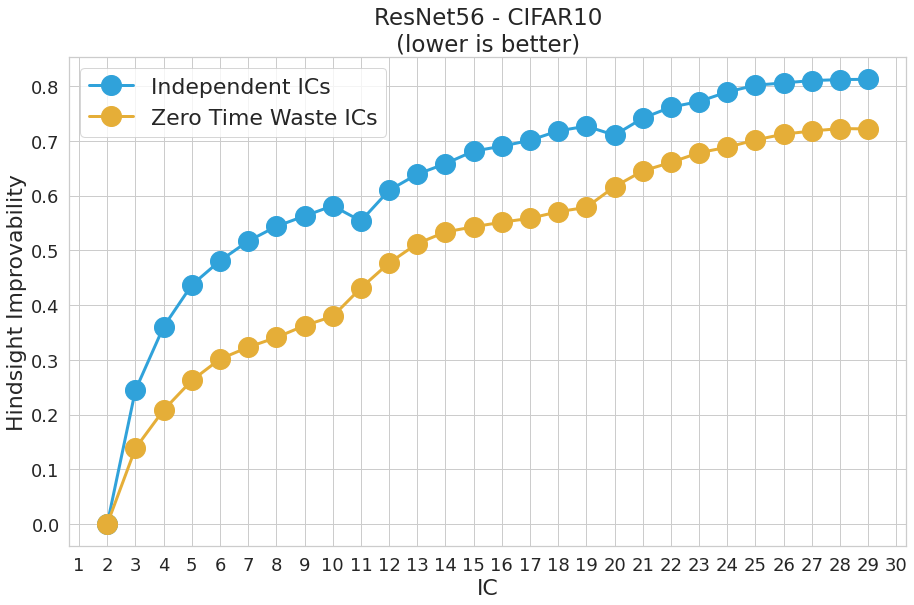}
    \end{subfigure}
    \begin{subfigure}[t]{0.48\textwidth}
        \includegraphics[width=\textwidth]{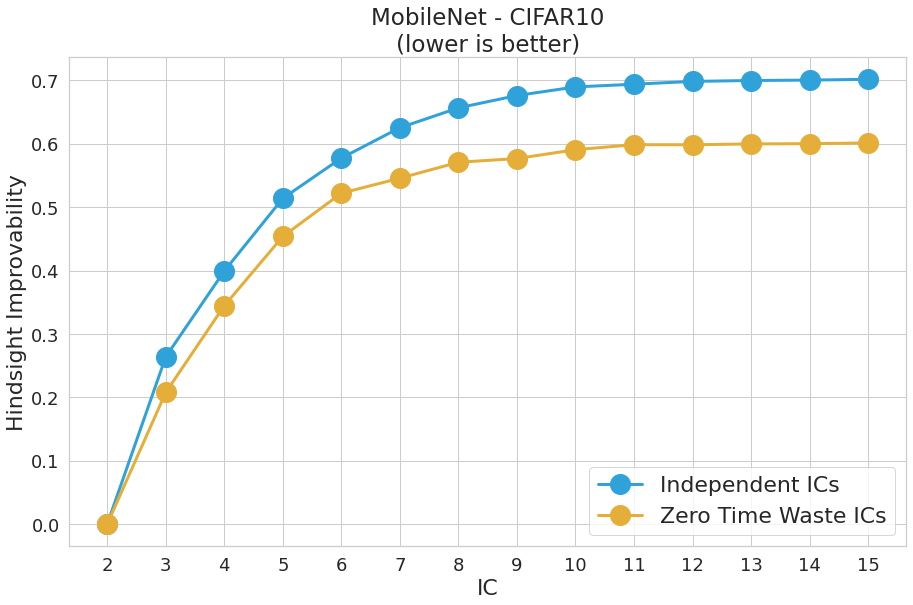}
    \end{subfigure}
    \begin{subfigure}[t]{0.48\textwidth}
        \includegraphics[width=\textwidth]{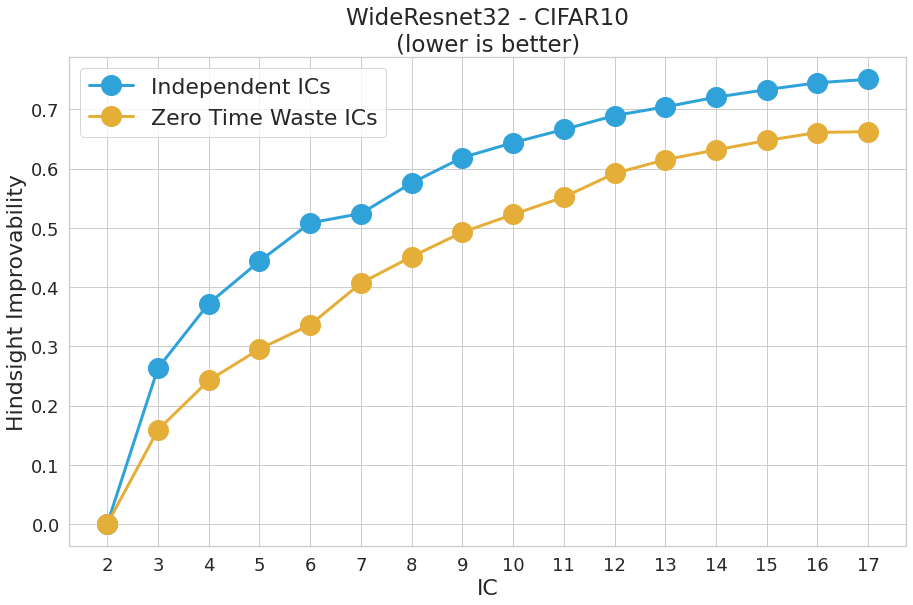}
    \end{subfigure}
    \begin{subfigure}[t]{0.48\textwidth}
        \includegraphics[width=\textwidth]{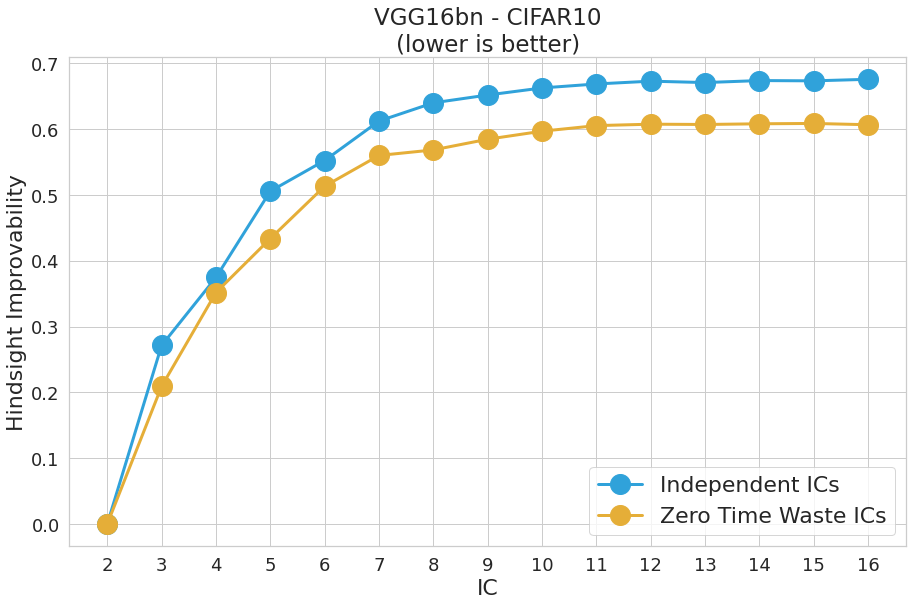}
    \end{subfigure}

    \caption{Hindsight Improvability of various architectures trained on CIFAR-10.}
    \label{fig:supp_hindsight_cifar10}
\end{figure*}

\begin{figure*}[h!]
    \centering
    \begin{subfigure}[t]{0.48\textwidth}
        \includegraphics[width=\textwidth]{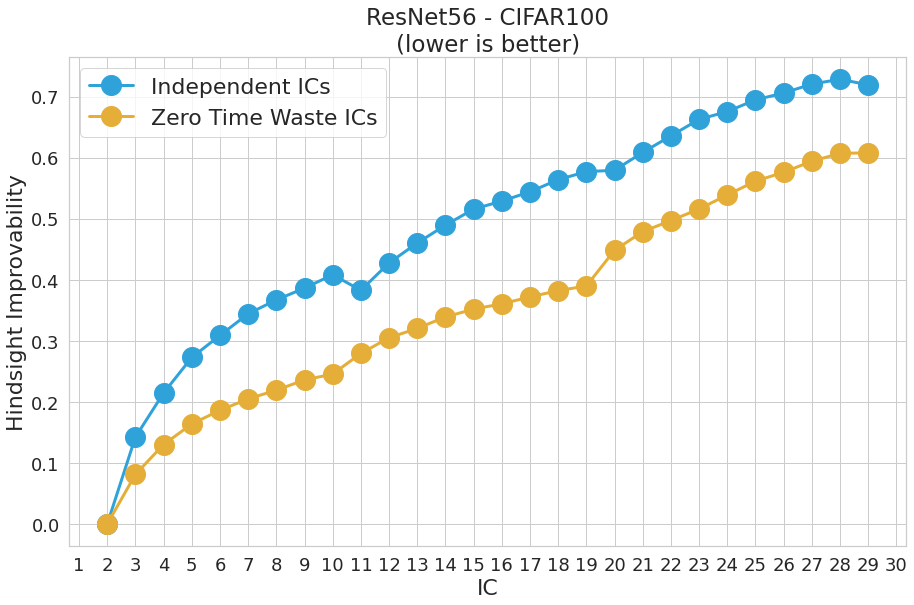}
    \end{subfigure}
    \begin{subfigure}[t]{0.48\textwidth}
        \includegraphics[width=\textwidth]{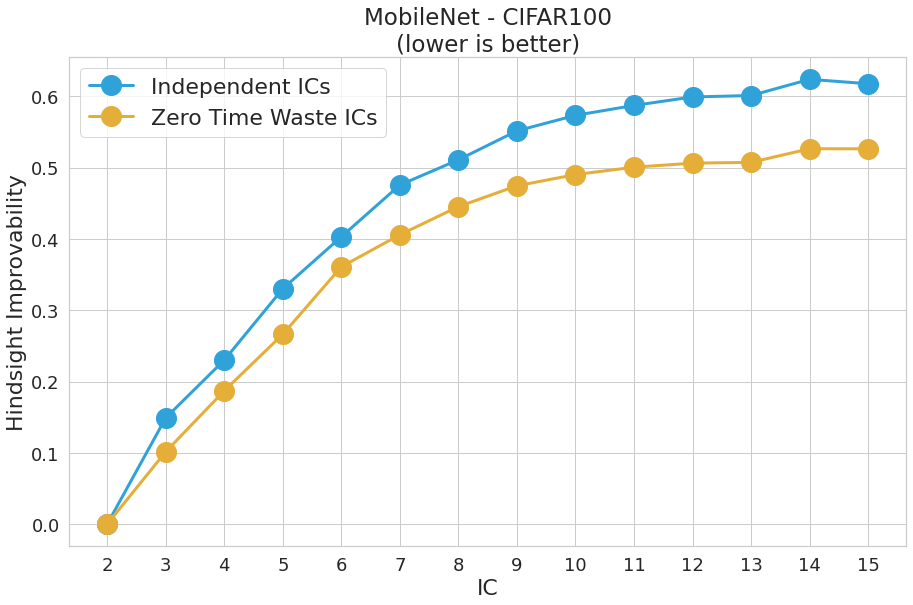}
    \end{subfigure}
    \begin{subfigure}[t]{0.48\textwidth}
        \includegraphics[width=\textwidth]{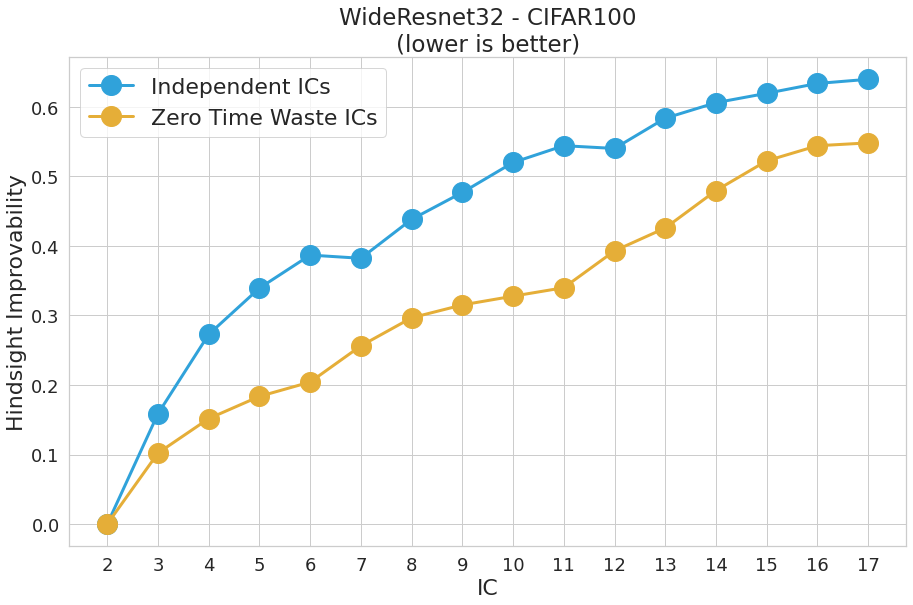}
    \end{subfigure}
    \begin{subfigure}[t]{0.48\textwidth}
        \includegraphics[width=\textwidth]{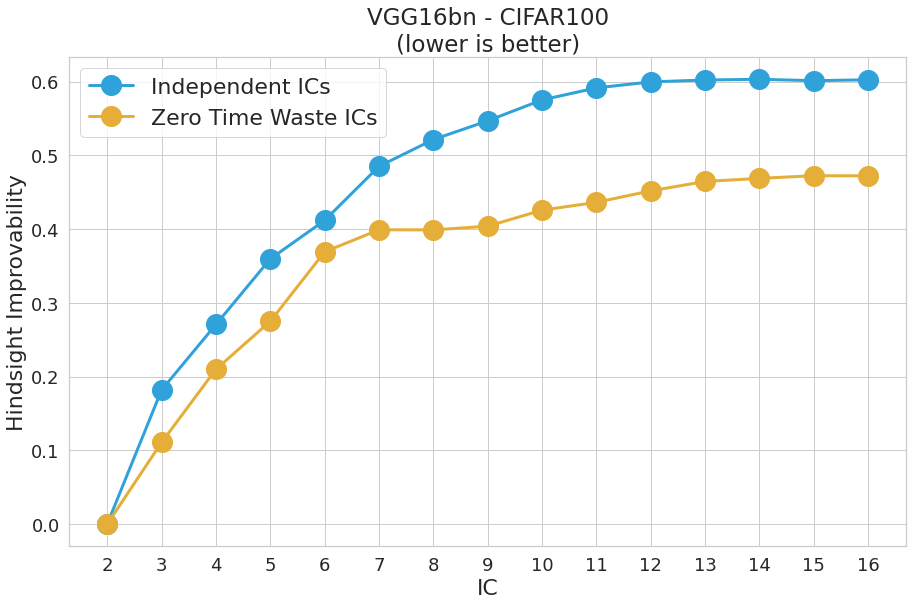}
    \end{subfigure}

    \caption{Hindsight Improvability of various architectures trained on CIFAR-100.}
    \label{fig:supp_hindsight_cifar100}
\end{figure*}

\begin{figure*}[h!]
    \centering
    \begin{subfigure}[t]{0.48\textwidth}
        \includegraphics[width=\textwidth]{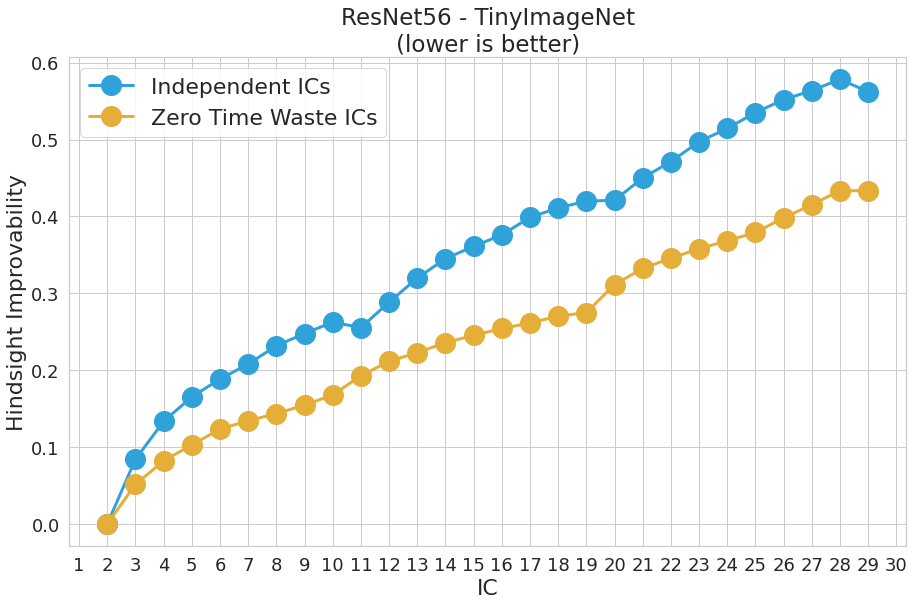}
    \end{subfigure}
    \begin{subfigure}[t]{0.48\textwidth}
        \includegraphics[width=\textwidth]{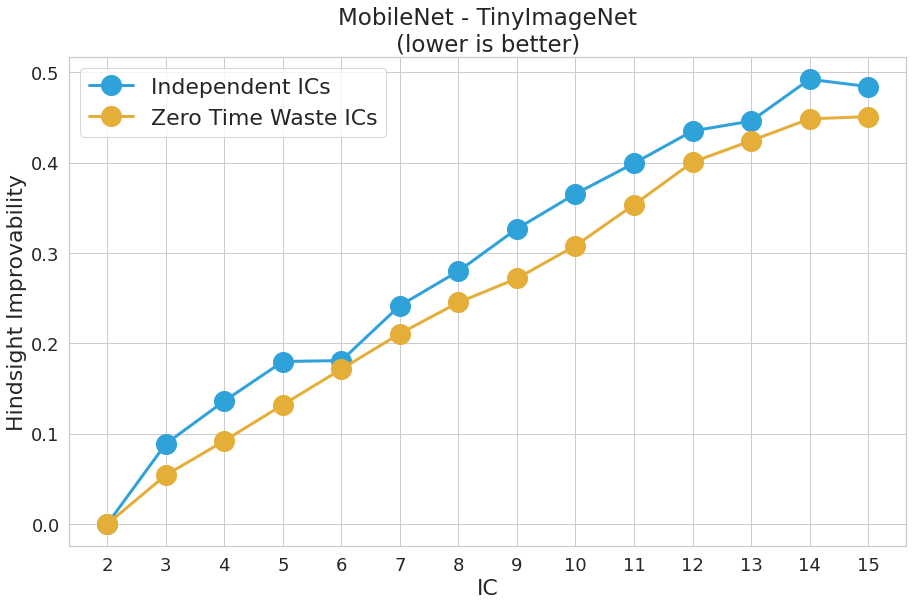}
    \end{subfigure}
    \begin{subfigure}[t]{0.48\textwidth}
        \includegraphics[width=\textwidth]{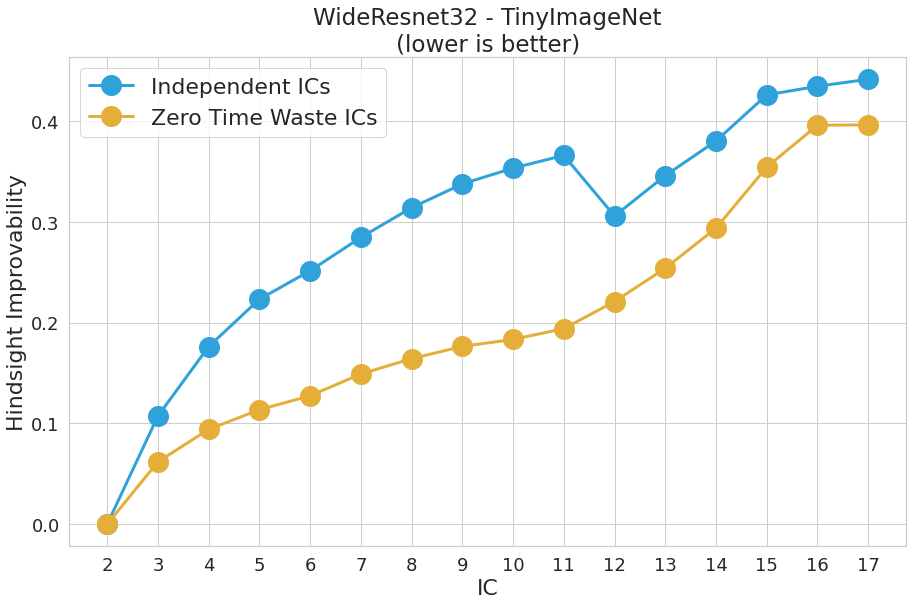}
    \end{subfigure}
    \begin{subfigure}[t]{0.48\textwidth}
        \includegraphics[width=\textwidth]{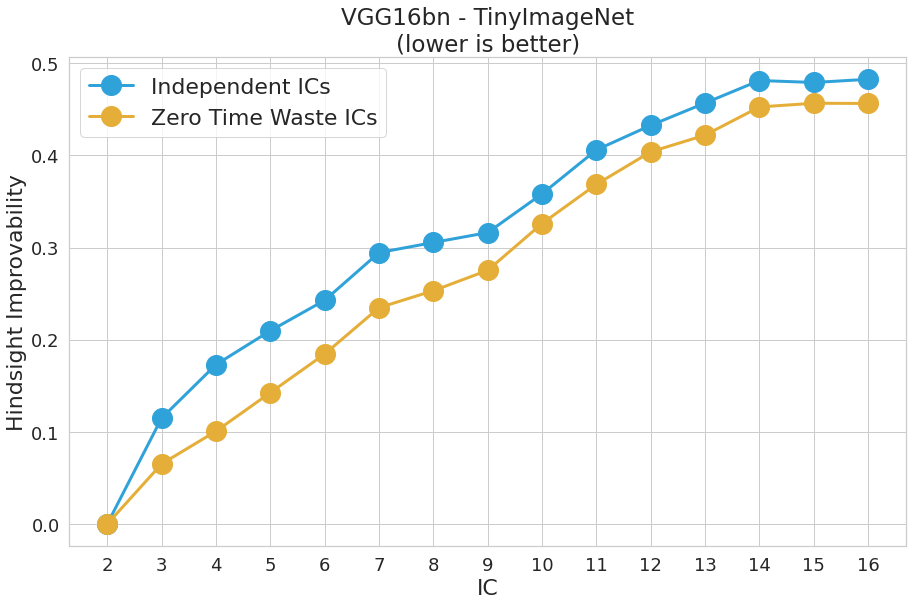}
    \end{subfigure}

    \caption{Hindsight Improvability of various architectures trained on Tiny~ImageNet.}
    \label{fig:supp_hindsight_tinyimagenet}
\end{figure*}


%% file: neurips_2021.bbl
\begin{thebibliography}{10}

\bibitem{arielynorton2011fromthinkingtoolittle}
Dan Ariely and Michael~I. Norton.
\newblock From thinking too little to thinking too much: a continuum of
  decision making.
\newblock {\em WIREs Cognitive Science}, 2(1):39--46, 2011.

\bibitem{bengio2013estimating}
Yoshua Bengio, Nicholas L{\'e}onard, and Aaron Courville.
\newblock Estimating or propagating gradients through stochastic neurons for
  conditional computation.
\newblock {\em arXiv:1308.3432}, 2013.

\bibitem{bentejac2020comparative}
Candice Bent{\'e}jac, Anna Cs{\"o}rg{\H{o}}, and Gonzalo
  Mart{\'\i}nez-Mu{\~n}oz.
\newblock A comparative analysis of gradient boosting algorithms.
\newblock {\em Artificial Intelligence Review}, pages 1--31, 2020.

\bibitem{berestizshevsky2019dynamically}
Konstantin Berestizshevsky and Guy Even.
\newblock Dynamically sacrificing accuracy for reduced computation: cascaded
  inference based on softmax confidence.
\newblock In {\em Proceedings of the International Conference on Artificial
  Neural Networks}, ICANN, pages 306--320. Springer, 2019.

\bibitem{davis2013low}
Andrew Davis and Itamar Arel.
\newblock Low-rank approximations for conditional feedforward computation in
  deep neural networks.
\newblock {\em arXiv:1312.4461}, 2013.

\bibitem{dietterich2000ensemble}
Thomas~G Dietterich.
\newblock Ensemble methods in machine learning.
\newblock In {\em Proceedings of the International Workshop on Multiple
  Classifier Systems}, page~15. Springer, 2000.

\bibitem{DBLP:journals/corr/abs-1904-12901}
Gabriel Dulac{-}Arnold, Daniel~J. Mankowitz, and Todd Hester.
\newblock Challenges of real-world reinforcement learning.
\newblock {\em CoRR}, abs/1904.12901, 2019.

\bibitem{fort2019deep}
Stanislav Fort, Huiyi Hu, and Balaji Lakshminarayanan.
\newblock Deep ensembles: a loss landscape perspective.
\newblock {\em arXiv:1912.02757}, 2019.

\bibitem{gigerenzer2011heuristicdecisionmaking}
Gerd Gigerenzer and Wolfgang Gaissmaier.
\newblock Heuristic decision making.
\newblock {\em {Annual Review of Psychology}}, 62:451--82, 2011.

\bibitem{grigorescu2020survey}
Sorin Grigorescu, Bogdan Trasnea, Tiberiu Cocias, and Gigel Macesanu.
\newblock A survey of deep learning techniques for autonomous driving.
\newblock {\em Journal of Field Robotics}, 37(3):362--386, 2020.

\bibitem{he2016deep}
Kaiming He, Xiangyu Zhang, Shaoqing Ren, and Jian Sun.
\newblock Deep residual learning for image recognition.
\newblock In {\em Proceedings of the IEEE Conference on Computer Vision and
  Pattern Recognition}, CVPR, pages 770--778, 2016.

\bibitem{he2017channel}
Yihui He, Xiangyu Zhang, and Jian Sun.
\newblock Channel pruning for accelerating very deep neural networks.
\newblock In {\em Proceedings of the IEEE International Conference on Computer
  Vision}, pages 1389--1397, 2017.

\bibitem{MLJ12-hester}
Todd Hester and Peter Stone.
\newblock {TEXPLORE}: Real-time sample-efficient reinforcement learning for
  robots.
\newblock {\em Machine Learning}, 90(3):385--429, 2013.

\bibitem{44873}
Geoffrey Hinton, Oriol Vinyals, and Jeffrey Dean.
\newblock Distilling the knowledge in a neural network.
\newblock In {\em Proceedings of the NIPS Workshop on Deep Learning and
  Representation Learning}, 2015.

\bibitem{howard2017mobilenets}
Andrew~G Howard, Menglong Zhu, Bo~Chen, Dmitry Kalenichenko, Weijun Wang,
  Tobias Weyand, Marco Andreetto, and Hartwig Adam.
\newblock Mobilenets: Efficient convolutional neural networks for mobile vision
  applications.
\newblock {\em arXiv:1704.04861}, 2017.

\bibitem{huang2018multi}
Gao {Huang}, Danlu {Chen}, Tianhong {Li}, Felix {Wu}, Laurens van~der {Maaten},
  and Kilian~Q. {Weinberger}.
\newblock Multi-scale dense networks for resource efficient image
  classification.
\newblock In {\em International Conference on Learning Representations}, 2018.

\bibitem{jung2018perception}
Sunggoo Jung, Sunyou Hwang, Heemin Shin, and David~Hyunchul Shim.
\newblock Perception, guidance, and navigation for indoor autonomous drone
  racing using deep learning.
\newblock {\em IEEE Robotics and Automation Letters}, 3(3):2539--2544, 2018.

\bibitem{kahneman2017thinking}
Daniel Kahneman.
\newblock {\em Thinking, fast and slow}.
\newblock Farrar, Straus and Giroux, 2017.

\bibitem{kaya2019shallow}
Yigitcan Kaya, Sanghyun Hong, and Tudor Dumitras.
\newblock Shallow-deep networks: Understanding and mitigating network
  overthinking.
\newblock In {\em Proceedings of the International Conference on Machine
  Learning}, ICML, pages 3301--3310, 2019.

\bibitem{kermany2018identifying}
Daniel~S Kermany, Michael Goldbaum, Wenjia Cai, Carolina~CS Valentim, Huiying
  Liang, Sally~L Baxter, Alex McKeown, Ge~Yang, Xiaokang Wu, Fangbing Yan,
  et~al.
\newblock Identifying medical diagnoses and treatable diseases by image-based
  deep learning.
\newblock {\em Cell}, 172(5):1122--1131, 2018.

\bibitem{kouris2019approximate}
Alexandros Kouris, Stylianos~I. Venieris, Michail Rizakis, and Christos-Savvas
  Bouganis.
\newblock Approximate {LSTM}s for time-constrained inference: Enabling fast
  reaction in self-driving cars, 2019.

\bibitem{krizhevsky2017imagenet}
Alex Krizhevsky, Ilya Sutskever, and Geoffrey~E Hinton.
\newblock Imagenet classification with deep convolutional neural networks.
\newblock {\em Communications of the ACM}, 60(6):84--90, 2017.

\bibitem{lakshminarayanan2017simple}
Balaji Lakshminarayanan, Alexander Pritzel, and Charles Blundell.
\newblock Simple and scalable predictive uncertainty estimation using deep
  ensembles.
\newblock In {\em Advances in Neural Information Processing Systems}, NIPS,
  pages 6402--6413, 2017.

\bibitem{lee2021gst}
Juhyoung Lee, Sangyeob Kim, Sangjin Kim, Wooyoung Jo, and Hoi-Jun Yoo.
\newblock Gst: Group-sparse training for accelerating deep reinforcement
  learning, 2021.

\bibitem{li2019improved}
Hao {Li}, Hong {Zhang}, Xiaojuan {Qi}, Yang {Ruigang}, and Gao {Huang}.
\newblock Improved techniques for training adaptive deep networks.
\newblock In {\em 2019 IEEE/CVF International Conference on Computer Vision
  (ICCV)}, pages 1891--1900, 2019.

\bibitem{livne2020pops}
Dor Livne and Kobi Cohen.
\newblock Pops: Policy pruning and shrinking for deep reinforcement learning,
  2020.

\bibitem{mcgill2017deciding}
Mason {McGill} and Pietro {Perona}.
\newblock Deciding how to decide: dynamic routing in artificial neural
  networks.
\newblock In {\em ICML'17 Proceedings of the 34th International Conference on
  Machine Learning - Volume 70}, pages 2363--2372, 2017.

\bibitem{mnih2015human}
Volodymyr Mnih, Koray Kavukcuoglu, David Silver, Andrei~A Rusu, Joel Veness,
  Marc~G Bellemare, Alex Graves, Martin Riedmiller, Andreas~K Fidjeland, Georg
  Ostrovski, et~al.
\newblock Human-level control through deep reinforcement learning.
\newblock {\em Nature}, 518(7540):529--533, 2015.

\bibitem{phuong2019distillation}
Mary {Phuong} and Christoph {Lampert}.
\newblock Distillation-based training for multi-exit architectures.
\newblock In {\em 2019 IEEE/CVF International Conference on Computer Vision
  (ICCV)}, pages 1355--1364, 2019.

\bibitem{stable-baselines3}
Antonin Raffin, Ashley Hill, Maximilian Ernestus, Adam Gleave, Anssi
  Kanervisto, and Noah Dormann.
\newblock Stable baselines3.
\newblock \url{https://github.com/DLR-RM/stable-baselines3}, 2019.

\bibitem{scardapane2020differentiable}
Simone Scardapane, Danilo Comminiello, Michele Scarpiniti, Enzo Baccarelli, and
  Aurelio Uncini.
\newblock Differentiable branching in deep networks for fast inference.
\newblock In {\em Proceedings of the IEEE International Conference on
  Acoustics, Speech and Signal Processing}, ICASSP, pages 4167--4171, 2020.

\bibitem{scardapane2020should}
Simone Scardapane, Michele Scarpiniti, Enzo Baccarelli, and Aurelio Uncini.
\newblock Why should we add early exits to neural networks?
\newblock {\em arXiv:2004.12814}, 2020.

\bibitem{schapire_strength_1990}
Robert~E. Schapire.
\newblock The strength of weak learnability.
\newblock {\em Machine Learning}, 5(2):197--227, 1990.

\bibitem{Schuitema2010ControlDI}
E.~Schuitema, L.~Bu\c{s}oniu, Robert Babu\v{s}ka, and P.~Jonker.
\newblock Control delay in reinforcement learning for real-time dynamic
  systems: A memoryless approach.
\newblock {\em 2010 IEEE/RSJ International Conference on Intelligent Robots and
  Systems}, pages 3226--3231, 2010.

\bibitem{schulman2015high}
John Schulman, Philipp Moritz, Sergey Levine, Michael Jordan, and Pieter
  Abbeel.
\newblock High-dimensional continuous control using generalized advantage
  estimation.
\newblock {\em arXiv:1506.02438}, 2015.

\bibitem{schulman2017proximal}
John Schulman, Filip Wolski, Prafulla Dhariwal, Alec Radford, and Oleg Klimov.
\newblock Proximal policy optimization algorithms.
\newblock {\em arXiv preprint arXiv:1707.06347}, 2017.

\bibitem{DBLP:journals/corr/SimonyanZ14a}
Karen Simonyan and Andrew Zisserman.
\newblock Very deep convolutional networks for large-scale image recognition.
\newblock In {\em Proceedings of the International Conference on Learning
  Representations}, ICLR, 2015.

\bibitem{teerapittayanon2016branchynet}
Surat Teerapittayanon, Bradley McDanel, and Hsiang-Tsung Kung.
\newblock Branchynet: Fast inference via early exiting from deep neural
  networks.
\newblock In {\em Proceedings of the International Conference on Pattern
  Recognition}, ICPR, pages 2464--2469, 2016.

\bibitem{viola2004robust}
Paul {Viola} and Michael~J. {Jones}.
\newblock Robust real-time face detection.
\newblock {\em International Journal of Computer Vision}, 57(2):137--154, 2004.

\bibitem{wang2017idk}
Xin {Wang}, Yujia {Luo}, Daniel {Crankshaw}, Alexey {Tumanov}, Fisher {Yu}, and
  Joseph~E. {Gonzalez}.
\newblock Idk cascades: Fast deep learning by learning not to overthink.
\newblock In {\em UAI}, pages 580--590, 2017.

\bibitem{wang2018skipnet}
Xin Wang, Fisher Yu, Zi-Yi Dou, Trevor Darrell, and Joseph~E Gonzalez.
\newblock Skipnet: learning dynamic routing in convolutional networks.
\newblock In {\em Proceedings of the European Conference on Computer Vision},
  ECCV, pages 409--424, 2018.

\bibitem{wang2020deep}
Xin Wang, Fisher Yu, Lisa Dunlap, Yi-An Ma, Ruth Wang, Azalia Mirhoseini,
  Trevor Darrell, and Joseph~E Gonzalez.
\newblock Deep mixture of experts via shallow embedding.
\newblock In {\em Proceedings of the Uncertainty in Artificial Intelligence},
  UAI, pages 552--562, 2020.

\bibitem{wang2020glance}
Yulin {Wang}, Kangchen {Lv}, Rui {Huang}, Shiji {Song}, Le~{Yang}, and Gao
  {Huang}.
\newblock Glance and focus: a dynamic approach to reducing spatial redundancy
  in image classification.
\newblock In {\em Advances in Neural Information Processing Systems},
  volume~33, pages 2432--2444, 2020.

\bibitem{yang2020resolution}
Le~{Yang}, Yizeng {Han}, Xi~{Chen}, Shiji {Song}, Jifeng {Dai}, and Gao
  {Huang}.
\newblock Resolution adaptive networks for efficient inference.
\newblock In {\em 2020 IEEE/CVF Conference on Computer Vision and Pattern
  Recognition (CVPR)}, pages 2369--2378, 2020.

\bibitem{zagoruyko2016wide}
Sergey Zagoruyko and Nikos Komodakis.
\newblock Wide residual networks.
\newblock {\em arXiv:1605.07146}, 2016.

\bibitem{8852451}
Hongjie Zhang, Zhuocheng He, and Jing Li.
\newblock Accelerating the deep reinforcement learning with neural network
  compression.
\newblock In {\em 2019 International Joint Conference on Neural Networks
  (IJCNN)}, pages 1--8, 2019.

\bibitem{zhou2020bert}
Wangchunshu Zhou, Canwen Xu, Tao Ge, Julian McAuley, Ke~Xu, and Furu Wei.
\newblock {BERT} loses patience: fast and robust inference with early exit.
\newblock {\em arXiv:2006.04152}, 2020.

\end{thebibliography}
